\pdfoutput=1
\documentclass[review]{fcs}

\usepackage{parskip} % 自动处理段落间距和缩进
\usepackage{setspace}
\usepackage{float}
\usepackage{tikz}
\usepackage[edges]{forest}

\usepackage{color}
\usepackage[normalem]{ulem}
\usepackage{tabularray}

\definecolor{hidden-draw}{RGB}{20,68,106}
\definecolor{hidden-pink}{RGB}{255,245,247}

\title{A Survey on LoRA of Large Language Models}
\author[1]{Yuren MAO} 
\author[1]{Yuhang GE} 
\author[1]{Yijiang FAN} 
\author[1]{Wenyi XU} 
\author[1]{Yu MI} 
\author[1]{\\Zhonghao HU} 
\author*[1]{Yunjun GAO} 
\address[1]{Zhejiang University, Hangzhou 310007, China}
\fcssetup{
  received       = {month dd, yyyy},
  accepted       = {month dd, yyyy},
  corr-email     = {gaoyj@zju.edu.cn},
}

\begin{abstract}
Low-Rank Adaptation~(LoRA), which updates the dense neural network layers with pluggable low-rank matrices, is one of the best performed parameter efficient fine-tuning paradigms. Furthermore, it has significant advantages in cross-task generalization and privacy-preserving. Hence, LoRA has gained much attention recently, and the number of related literature demonstrates exponential growth. It is necessary to conduct a comprehensive overview of the current progress on LoRA. This survey categorizes and reviews the progress from the perspectives of (1) downstream adaptation improving variants that improve LoRA's performance on downstream tasks; (2) cross-task generalization methods that mix multiple LoRA plugins to achieve cross-task generalization; (3) efficiency-improving methods that boost the computation-efficiency of LoRA; (4) data privacy-preserving methods that use LoRA in federated learning; (5) application. Besides, this survey also discusses the future directions in this field. At last, we provide a Github page~\footnote{\href{https://github.com/ZJU-LLMs/Awesome-LoRAs.git}{https://github.com/ZJU-LLMs/Awesome-LoRAs.git}} for readers to check the updates and initiate discussions on this survey paper.
\end{abstract}

\keywords{Low-Rank Adaptation, LoRA, Large Language Models, LLMs}

\begin{document}
\begin{sloppypar}

\section{Introduction}
Rapidly increasing parameter scales of pre-training language models improves their generalization ability and brings emergent abilities. In the last few years,  the parameter scales of pre-training languages models have increased by thousands of times (e.g., from 330M parameter BERT \cite{conf/naacl/DevlinCLT19} to 540B parameter PaLM \cite{journals/jmlr/palm23}). These pre-training language models having large parameter scales are termed Large language models (LLMs). Nevertheless, due to the knowledge boundaries of the LLMs, their abilities on some downstream tasks are still limited. To expand the knowledge boundaries, it remains necessary to fine-tune LLMs on the downstream tasks.

However, fine-tuning the full parameters of an LLM, namely full fine-tuning, is extremely computationally expensive, for example, full fine-tuning of a LLaMA2-7B ~\cite{journals/corr/abs-2309-12307} model requires approximately 60GB of memory, which exceeds the capacity of common consumer GPUs \cite{journals/corr/abs-2403-17919}.
 To reduce the computational cost, various parameter-efficient fine-tuning (PEFT) methods have been proposed \cite{journals/natmi/Ding23}. They adapt LLMs to downstream tasks by only fine-tuning a small number of (extra) model parameters. From the perspective of whether extra parameters are involved, PEFT methods can be divided into two categories: extra-parameter methods and intra-parameter methods. The extra-parameter methods freeze all of the original parameters of an LLM and insert a set of learnable parameters to optimize the model input or model layers such as adapter tuning \cite{conf/icml/adapter} and prompt tuning \cite{conf/emnlp/LesterAC21}. By contrast, intra-parameter methods freeze most of the original parameters of an LLM and only tune a small number of parameters of the LLM such as BitFit \cite{conf/acl/ZakenGR22}, LISA~\cite{journals/corr/abs-2403-17919} and LoRA \cite{conf/iclr/HuSWALWWC22}.

When we do not have access to modify the model architecture, intra-parameter methods are desirable. Among the intra-parameter methods, LoRA is the most widely used one, because it can achieve a comparable or better downstream adaptation performance to the full fine-tuning on a range of downstream tasks \cite{conf/iclr/HuSWALWWC22} and is easy to implement. Besides, there are many variants have been proposed to further improve the downstream adaptation ability of LoRA on more challenging downstream tasks.

LoRA achieves parameter efficiency by updating the dense neural network layers of an LLM with pluggable low-rank matrices. These matrices (a.k.a, LoRA plugins) are independent of the LLM, which can be stored and reused in other related downstream tasks. Furthermore, these LoRA plugins can be combined to achieve cross-task generalization, which can facilitate multi-task learning, domain adaptation, and continual learning.
% for LLMs.
% This pluggable character enables LoRA and its varaints to achieve cross-task generalization.

As the LoRA modules accumulate, the computation cost of managing LoRA modules is increasing. Although LoRA is computation-efficient, the computational cost of managing a larger number of LoRA modules is unignorable. It is necessary to further improve the computation efficiency of LoRA. The improvement can come from reducing the computation cost of single LoRA modules and accelerating the scalable serving of multiple modules. It can boost the application of LoRA in real-world use cases, such as Generative-as-a-Service (GaaS) cloud products. 

In some cases, the training data are privately owned by multiple clients and cannot be centralized. To adapt LLMs with the distributed training data, we can adopt federated learning to protect the data privacy of each client. However, federated learning suffers expensive communication and computation costs. To reduce costs, LoRA is a natural choice. Its parameter-efficient nature helps to reduce the computation cost of each client and the communication cost of sharing parameters across clients. %Furthermore, the pluggable feature of LoRA can help preserve the parameter privacy of each client in federated learning. 
Furthermore, the pluggable feature of LoRA, which supports the localization or encryption of personalized parameters, enhances privacy protection within federated learning.
% Additionally, the design flexibility of LoRA allows it to address issues of data, device and model heterogeneity, facilitating a more uniform and effective learning process across diverse environments and platforms. 
Therefore, LoRA has a great potential for privacy-preserving. 

While some previous surveys have mentioned LoRA~\cite{journals/natmi/Ding23,zhao2023survey,han2024parameter}, they mainly focus on PEFT and only introduce a small number of LoRA-related works, lacking  systematic treatment and comprehensive overview on LoRA and its variants.
% lack comprehensive coverage of its variants. Additionally, they do not systematically introduce cross-task generalization, efficiency improvements, LoRA for federated learning, and LoRA-based applications, hindering the ability to keep pace with the rapid development of LoRA.}
% 
In this survey, we give a comprehensive overview of the current progress on LoRA for methods (1) improving downstream adaption performance of LoRA; (2) mixing LoRA modules to achieve cross-task generalization; (3) boosting the computation-efficiency of LoRA; (4) adopting LoRA in federated learning. Besides, the application of LoRA is briefly introduced.  This taxonomy of LoRA-related methods is illustrated in Figure~\ref{fig:taxonomy}. This survey is expected to give comprehensive background knowledge, research trends and technical insights for LoRA.

The rest of this survey is organized as follows. Section~\ref{sec:LoRA} introduces the background knowledge of LoRA, and Section~\ref{sec:Downstream_Adaptation} introduces the LoRA's variants that aim to improve the downstream adaptation performance. In Section~\ref{sec:Cross-task_Generalization}, we review the LoRA mixture methods that mix LoRA modules to achieve cross-task generalization. The LoRA-driven federated learning methods are introduced in Section~\ref{sec:LoRA_Federate_Learning}. 
Section~\ref{sec:Applications} reports the applications of LoRA.
We conclude this survey and discuss the future directions in Section~\ref{sec:Conclusion}.

% \input{Sections/framework.tex}
% \newpage

\section{Low-Rank Adaptation (LoRA)}
\label{sec:LoRA}
\renewcommand{\dblfloatpagefraction}{0.9}
\tikzstyle{my-box}=[
    rectangle,
    draw=hidden-draw,
    rounded corners,
    text opacity=1,
    minimum height=6em,
    minimum width=5em,
    inner sep=2pt,
    align=center,
    fill opacity=.5,
    line width=1.0pt,
]   
\tikzstyle{leaf}=[my-box, minimum height=2.8em,
    fill=hidden-pink!80, text=black, align=left,font=\normalsize,
    inner xsep=2pt,
    inner ysep=4pt,
    line width=1.2pt,
]

\begin{figure*}[!ht]
    \vspace{5mm}
    \centering
    \resizebox{\textwidth}{!}{
        \begin{forest}
            forked edges,
            for tree={
                grow=east,
                reversed=true,
                anchor=base west,
                parent anchor=east,
                child anchor=west,
                base=center,
                font=\large,
                rectangle,
                draw=hidden-draw,
                rounded corners,
                align=left,
                text centered,
                minimum width=4em,
                edge+={darkgray, line width=1pt},
                s sep=3pt,
                inner xsep=2pt,
                inner ysep=3pt,
                line width=0.8pt,
                ver/.style={rotate=90, child anchor=north, parent anchor=south, anchor=center},
            },
            where level=1{text width=16em,font=\normalsize,}{},
            where level=2{text width=18em,font=\normalsize,}{},
            where level=3{text width=20em,font=\normalsize,}{},
            where level=4{text width=20em,font=\normalsize,}{},
            where level=5{text width=20em,font=\normalsize,}{},
            [
                \textbf{Low-Rank Adaptation of Large Language Models}, ver        
                [
                    \textbf{Low-Rank Adaptation}(\S\ref{sec:LoRA}), fill=magenta!10
                    [
                        \textbf{Theoretical Analysis}(\S\ref{subsec:theoretical_analysis}), fill=magenta!10
                        [
                            \textbf{Malladi et al.}~\cite{conf/icml/MalladiWYCA23}{,}
                            \textbf{Koubbi et al.}~\cite{journals/corr/abs-2402-15415}{,}
                            \textbf{Jang et al.}~\cite{journals/corr/abs-2402-11867}{,}\\
                            \textbf{Zhu et al.}~\cite{journals/corr/abs-2402-16842}{,}
                            \textbf{Zeng et al.}~\cite{journals/corr/abs-2310-17513}, leaf, text width=20em
                        ]
                    ] 
                    [
                        \textbf{Beyond Fine-tuning}(\S\ref{subsec:beyond_ft}), fill=magenta!10
                        [
                            \textbf{ReLoRA}~\cite{lialin2023relora}{,}
                            \textbf{MoRA}~\cite{jiang2024mora}{,}
                            \textbf{LTE}~\cite{journals/corr/abs-2402-16828}{,}
                            \textbf{InfLoRA}~\cite{journals/corr/abs-2404-00228}{,}\\
                            \textbf{GS-LoRA}~\cite{journals/corr/abs-2403-11530}{,}
                            \textbf{I-LoRA}~\cite{ren2024analyzing}{,}
                            \textbf{LongLoRA}~\cite{journals/corr/abs-2309-12307}{,}\\
                            \textbf{SinkLoRA}~\cite{journals/corr/zhang2024sinklora}, leaf, text width=20em
                        ] 
                    ] 
                ] 
                [
                        \textbf{Downstream Adaptation Improving}(\S\ref{sec:Downstream_Adaptation}), fill=green!10
                        [
                            \textbf{Breaking the Low-rank Bottleneck}(\S\ref{subsec:Breaking_Bottleneck}), fill=green!10
                            [
                                \textbf{Stacking LoRAs along Fine-tuning}(\S\ref{subsubsec:Stacking_Along_Ft}),
                                [
                                    \textbf{ReLoRA}~\cite{lialin2023relora}{,}
                                    \textbf{COLA}~\cite{journals/corr/abs-2401-04151}{,}
                                    \textbf{MELoRA}~\cite{journals/corr/abs-2402-17263}, leaf, text width=18em
                                ]
                            ]
                            [
                                \textbf{Updating as gradient compressor}(\S\ref{subsubsec:Avoiding_Gradient_Compression}),
                                [
                                \textbf{FLoRA}~\cite{journals/corr/abs-2402-03293}, leaf, text width=18em
                                ]
                            ]
                            [
                                \textbf{Co-learning LLM and LoRA}(\S\ref{subsubsec:Co-updating_LLM_and_LoRA}),
                                [
                                \textbf{Delta-LoRA}~\cite{journals/corr/abs-2309-02411}, 
                                leaf, text width=18em
                                ]
                            ]                            
                        ]
                        [
                            \textbf{Dynamic Rank Allocation}(\S\ref{subsec:dynamic_rank_allocation}), fill=green!10
                            [
                                \textbf{SVD-Based Methods}(\S\ref{subsubsec:svd_based_methods}),
                                [
                                    \textbf{AdaLoRA}~\cite{conf/iclr/ZhangCBH0CZ23}{,}
                                    \textbf{SaLoRA}~\cite{Hu2023Structure}{,}
                                    \textbf{IncreLoRA}~\cite{journals/corr/abs-2308-12043}, leaf, text width=18em
                                ]
                            ]
                            [
                                \textbf{SRD-based Methods}(\S\ref{subsubsec:srd_based_methods}),
                                [
                                \textbf{DoRA~(Dynamic Low-Rank Adaptation)}~\cite{mao2024dora}{,} \\
                                \textbf{AutoLoRA}~\cite{journals/corr/abs-2403-09113}{,}
                                \textbf{SoRA}~\cite{conf/emnlp/DingLWCZL023}{,}
                                \textbf{ALoRA}~\cite{journals/corr/abs-2403-16187}, leaf, text width=18em
                                ]
                            ]
                            [
                                \textbf{Rank Sampling-based Methods}(\S\ref{subsubsec:rank_sampling_based_methods}),
                                [
                                \textbf{DyLoRA}~\cite{valipour2022dylora}, leaf, text width=18em
                                ]
                            ]                            
                        ]
                        [
                            \textbf{Optimizing the Learning Procedure}(\S\ref{subsec:opt_learning_procedure}), fill=green!10
                            [
                                \textbf{Initialization Improvement}(\S\ref{subsubsec:init_improvement}),
                                [
                                \textbf{Hayou et al.}~\cite{hayou2024impact}{,}
                                \textbf{PiSSA}~\cite{PiSSA}{,}
                                \textbf{MiLoRA}~\cite{MiLoRA}, leaf, text width=18em
                                ]
                            ]
                            [
                                \textbf{Gradient Update Optimization}(\S\ref{subsubsec:gradient_update_opt}),
                                [
                                \textbf{Zhang et al.}~\cite{zhang2024riemannian}{,}
                                \textbf{LoRA+}~\cite{lora+}{,}
                                \textbf{ResLoRA}~\cite{ResLoRA}{,}\\
                                \textbf{SIBO}\cite{SIBO}{,}
                                \textbf{Jin et al.}~\cite{derivative-free}{,} 
                                \textbf{DoRA}~\cite{DoRA}, leaf, text width=18em
                                ]
                            ]
                            [
                                \textbf{Overfitting Mitigation}(\S\ref{subsubsec:overfitting_mitigation}),
                                [
                                \textbf{BiLoRA}~\cite{BiLoRA}{,}
                                \textbf{Lin et al.}~\cite{lora-dropout}{,}
                                \textbf{HiddenKey}~\cite{HiddenKey}, leaf, text width=18em
                                ]
                            ]                        
                        ]                        
                        [
                            \textbf{Combining with other Learning Paradigms}(\S\ref{subsec:combining_with_other_learning_paradigms}), fill=green!10
                              [
                                \textbf{Laplace-LoRA}~\cite{journals/corr/abs-2308-13111}{,}
                                \textbf{PILLOW}~\cite{conf/emnlp/QiTSQXQ23}{,}
                                \textbf{STAR}~\cite{journals/corr/abs-2403-01165}, leaf, text width=20em
                              ]
                        ]                        
                    ]
                [
                        \textbf{Cross-task Generalization}(\S\ref{sec:Cross-task_Generalization}), fill=cyan!10
                        [
                            \textbf{Mixture with Manually Designed Weights}(\S\ref{subsec:Mixture with Manually Designed Weights}), fill=cyan!10
                            [
                            \textbf{Wang et al.}\cite{lora-ensembles}{,}
                            \textbf{Zhao et al.}\cite{lora-retriever}{,}
                            \textbf{Smith et al.}\cite{construct-vl}{,}\\
                            \textbf{ControlPE}\cite{sun2023or}{,}
                            \textbf{Zhang et al.}\cite{composing}{,}
                            \textbf{Chitale et al.}\cite{task-arithmetic}{,}\\
                            \textbf{Token-level Adaptation}\cite{token-level}{,}
                            \textbf{BYOM}\cite{byom},
                            leaf, text width=20em
                            ]
                        ]
                        [
                            \textbf{Mixture with Learnt Weights}(\S\ref{subsec:Mixture with Learnt Weights}), fill=cyan!10
                              [
                              \textbf{Asadi et al.}\cite{does}{,}
                               \textbf{LoRAHub}\cite{lorahub}{,}
                               \textbf{ComPEFT}\cite{compeft}{,}\\
                               \textbf{L-LoRA}\cite{l-lora}{,}
                               \textbf{MixLoRA}\cite{multimodal}{,}
                               \textbf{X-LoRA}\cite{x-lora}, leaf, text width=20em
                              ]
                        ]
                        [
                            \textbf{Mixture of LoRA Experts}(\S\ref{subsec:Mixture of LoRA Experts}), fill=cyan!10
                              [
                              \textbf{MoRAL}\cite{moral}{,}
                              \textbf{LoRAMoE}\cite{loramoe}{,}
                              \textbf{MoCLE}\cite{mocle}{,}\\
                              \textbf{MOELoRA}\cite{moelora}{,}
                              \textbf{Mixture-of-LoRAs}\cite{mixture-of-loras}{,}\\
                              \textbf{MultiLoRA}\cite{multilora}{,}
                              \textbf{MLoRE}\cite{multi-task}{,}
                              \textbf{MTLoRA}\cite{mtlora}{,}\\
                              \textbf{MoLA}\cite{higher}{,}
                              \textbf{LLaVA-MoLE}\cite{llava-mole}{,}
                              \textbf{SiRA}\cite{sira}{,}\\
                              \textbf{Octavius}\cite{octavius}{,}
                              \textbf{Fast LoRA}\cite{FLoRA}{,}
                              \textbf{MoSLoRA}\cite{wu2024mixture},
                                leaf, text width=20em
                              ]
                        ]                        
                    ]
                [
                        \textbf{Efficiency Improving}(\S\ref{sec:Computational_Efficiency}), fill=blue!10
                        [
                            \textbf{Parameter Reduction}(\S\ref{subsubsec:parameter-reduction}), fill=blue!10
                            [
                                \textbf{Parameter Freezing}(\S\ref{subsubsec:parameter-freezing}),
                                [
                                \textbf{LoRA-SP}\cite{lora-sp}{,}
                                \textbf{LoRA-FA}\cite{lora-fa}{,}
                                \textbf{AFLoRA}\cite{aflora}{,}\\
                                \textbf{DropBP}\cite{dropbp}{,}
                                \textbf{LoRA-XS}\cite{lora-xs}{,}
                                \textbf{BYOM-LoRA}\cite{byom},leaf, text width=18em
                                ]
                            ]
                            [
                                \textbf{Parameter Pruning}(\S\ref{subsubsec:parameter-pruning})
                                [
                                \textbf{LoRA-drop}\cite{lora-drop}{,}
                                \textbf{LoRAprune}\cite{loraprune}{,}\\
                                \textbf{LoRAshear}\cite{lorashear}{,}
                                \textbf{Zhu et al.}\cite{zhu2023parameter}, leaf, text width=18em
                                ]
                            ]
                            [
                                \textbf{Parameter Sharing}
                                (\S\ref{subsubsec:parameter-sharing})
                                [
                                \textbf{VeRA}\cite{vera}{,}
                                \textbf{VB-LoRA}\cite{vb-lora}{,}
                                \textbf{FourierFT}~\cite{gao2024parameter},
                                leaf, text width=18em
                                ]
                            ]
                        ]
                        [
                            \textbf{Parameter Quantization}(\S\ref{subsec:parameter-quantization}), fill=blue!10
                            [
                                \textbf{PTQ-based methods}
                                (\S\ref{subsubsec:ptq-methods})
                                [
                                \textbf{QLoRA}\cite{QLoRA}{,}
                                \textbf{QA-LoRA}\cite{QA-LoRA}, leaf, text width=18em
                                ]
                            ]
                            [
                                \textbf{QAT-base}
                                (\S\ref{subsubsec:qat-methods})       [
                                \textbf{LoftQ}\cite{LoftQ}{,}
                                \textbf{ApiQ}\cite{ApiQ}{,}
                                \textbf{L4Q}\cite{L4q}, leaf, text width=18em
                                ]
                            ]
                        ]
                        [
                            \textbf{Parallel LoRA Computing Frameworks}(\S\ref{subsec:parallel_lora_computing}), fill=blue!10
                            [
                                \textbf{Parallel Fine-tuning}(\S\ref{subsubsec:parallel_ft})
                                [
                                \textbf{ASPEN}\cite{aspen}, leaf, text width=18em
                                ]
                            ]
                            [
                                \textbf{Parallel Inference}(\S\ref{subsubsec:parallel-inference})
                                [
                                \textbf{Punica}\cite{punica}{,}
                                \textbf{S-LoRA}\cite{s-lora}{,}
                                \textbf{CARASERVE}\cite{caraserve}, leaf, text width=18em
                                ]
                            ]
                        ]
                    ]                
                    [
                    \textbf{LoRA for Federate Learning}(\S\ref{sec:LoRA_Federate_Learning}), fill=orange!10
                    [
                        \textbf{Data Heterogeneity}(\S\ref{subsec:data_hetero}), fill=orange!10
                        [
                        \textbf{SLoRA}~\cite{slora}{,}
                        \textbf{FeDeRA}~\cite{FeDeRA}{,}
                        \textbf{FFA-LoRA}~\cite{FFA-LoRA}, leaf, text width=20em
                        ]
                    ]
                    [
                        \textbf{Device Heterogeneity}(\S\ref{subsec:device_hetero}), fill=orange!10
                          [
                          \textbf{FedMS}~\cite{FedMS}{,}
                          \textbf{FlexLoRA}~\cite{FlexLoRA}{,}
                          \textbf{HETLORA}~\cite{HETLORA}, leaf, text width=20em
                          ]
                    ]
                    [
                        \textbf{Model Heterogeneity}(\S\ref{subsec:model_hetero}), fill=orange!10
                          [
                          \textbf{pFedLoRA}~\cite{pFedLoRA}, leaf, text width=20em
                          ]
                    ]                                        
                    [
                        \textbf{Parameter Privacy}(\S\ref{subsec:param_privacy}), fill=orange!10
                          [
                          \textbf{Huang et al.}~\cite{TEE}{,}
                          \textbf{PrivateLoRA}~\cite{PrivateLoRA}, leaf, text width=20em
                          ]
                    ]                             
                ]                
                [
                    \textbf{Applications of LoRA}(\S\ref{sec:Applications}), fill=yellow!10
                    [
                        \textbf{Language Tasks}(\S\ref{subsec:language_task}), fill=yellow!10
                        [
                        \textbf{Traditional NLP Task}\cite{zhang2024dialoguellm, li2023label, Bornheim_2024, lilong2024autore, alves2023steering, zheng2024fine, mujadia2023assessing, zhang2024personalized, liu2024tuning}{,}
                        \textbf{Code Task}\cite{liu2024delving, guo2024empirical, ayupov2022parameter, silva2023repairllama, roberson2024analyzing, pan2023stelocoder}{,}\\
                        \textbf{Model Alignment Task}\cite{sidahmed2024perl, santacroce2023efficient, sun2023exploring, quan2024dmoerm, zhang2024improving, zhai2023uncertaintypenalized, yang2024bayesian, yang2023bayesian}{,}\\
                        \textbf{Vertical Domain Task}\cite{tran2024bioinstruct, gema2024parameterefficient, toma2023clinical, suri2023suryakiran, ji2024assertion, wang2023ivygpt, bhatti2023sm70, konstantinidis2024finllama, pavlyshenko2023financial, liu2023fingpt, li2024ra,zhou2024db}, leaf, text width=20em
                        ]
                    ]
                    [
                        \textbf{Vision Task}(\S\ref{subsec:vision_task}), fill=yellow!10
                        [
                        \textbf{Image Generation Tasks}\cite{li2024diffstyler, frenkel2024implicit, liu2023facechain, liao2023calliffusion, shrestha2023style, li2024blockwise, kong2024omg, shi2023space, jin2023generating, wang2023customizing, guo2023smooth, cheng2024resadapter, smith2024continual, sun2023dreamsync, wang2023styleadapter, mix-of-show, luo2023lcmlora, golnari2023loraenhanced, ren2024customizeavideo, deng2024dragvideo, yang2023rerender, khandelwal2023infusion, blattmann2023stable, guo2023animatediff, huang2024dreamcontrol, ma2023xdreamer, yu2023boosting3d, yoo2024plausible, zhang2024dragtex}{,}\\
                        \textbf{Image Segmentation Task}\cite{ding2024samlp, ye2024sambased, na2024segment, chen2024samocta, feng2023cheap, zhang2023customized, wang2023sam, lin2024tracking, kong2023enhancing}, leaf, text width=20em
                        ]                    
                    ]
                    [
                        \textbf{Multimodal Tasks}(\S\ref{subsec:vision_task}), fill=yellow!10
                        [
                        \textbf{Audio-Text}\cite{chen2023salm}{,}
                        \textbf{Image-Text}\cite{dong2024internlmxcomposer2, ye2024mplugowl, lee2024collavo}{,}\\
                        \textbf{Video-Text}\cite{yeo2024visual, liu2024molca, ren2024tpllm}, leaf, text width=20em
                        ]
                    ] 
                ] 
            ]
            ]             
        \end{forest}
    }
    \vspace{-0mm}
    \caption{The taxonomy of this paper.}
    \label{fig:taxonomy}
    \vspace{5mm}
\end{figure*}
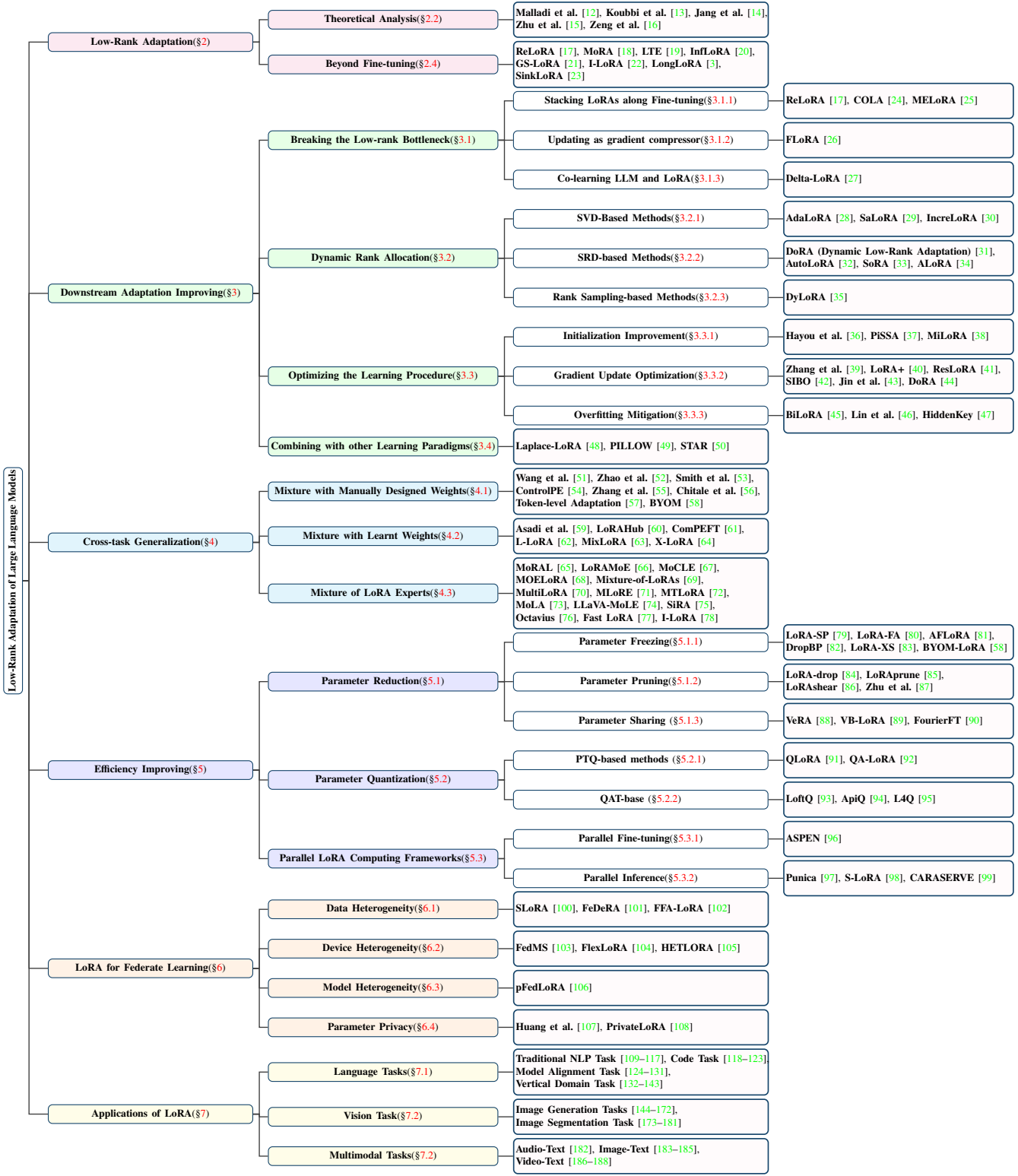

The Low-dimensional intrinsic dimensionality hypothesis~\cite{conf/acl/AghajanyanGZ20}
presents that over-parameterized models reside on a low intrinsic dimension, which demonstrates that we can achieve proper learning performance by only updating parameters related to the intrinsic rank. 
Based on this hypothesis, LoRA~\cite{conf/iclr/HuSWALWWC22} proposes to update dense layers in a model with low-rank matrices. It can achieve both parameter- and computational- efficiency. 
In this section, we first introduce the details of LoRA and then introduce existing works that focus on the theoretical analysis of LoRA. Furthermore, we demonstrate LoRA's efficiency in practice. At last, this section presents that LoRA can be used in other use cases except fine-tuning.

\subsection{LoRA}
\begin{figure*}[!ht]
    \centering
    \includegraphics[width=1.0\linewidth]{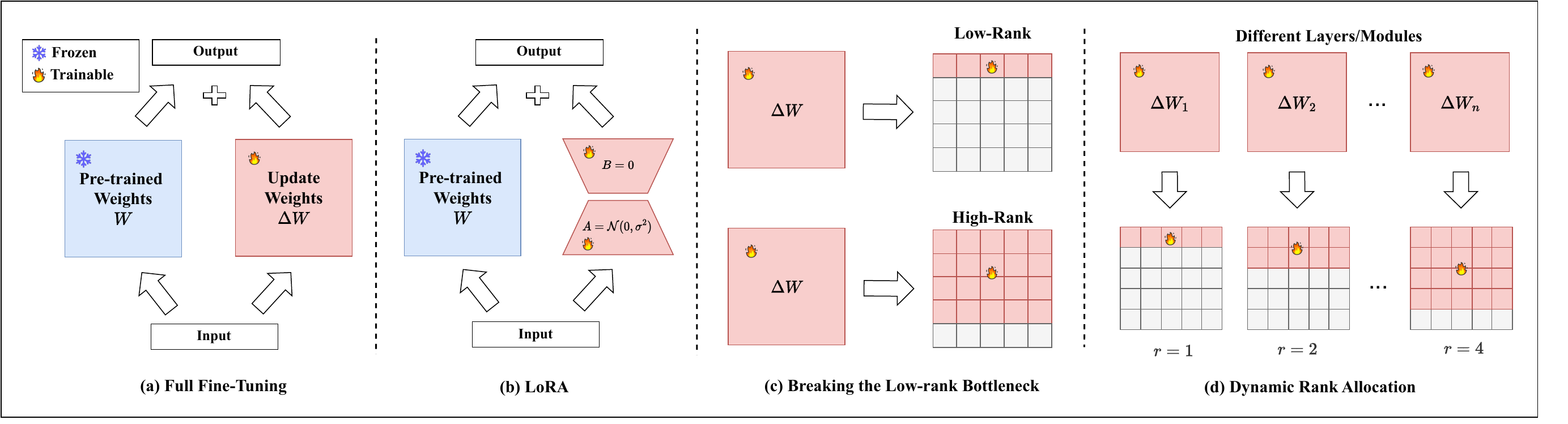}
    \caption{An illustration of full fine-tuning~(a), LoRA~(b) and its variants for improving downstream adaptation, which includes breaking the low-rank bottleneck~(c) and dynamic rank allocation~(d).}
    \label{fig:lora}
\end{figure*}

Given a dense neural network layer parameterized by $W_0 \in \mathbb{R}^{d \times k}$, to adapt it to a downstream task, we update it with $\Delta{W} \in \mathbb{R}^{d \times k}$ and obtain an updated layer parameterized by $W = W_0 + \Delta{W}$. For full fine-tuning, $\Delta{W}$ is computed based on gradients of all the $d \times k$ parameters for the layer, which is computationally expensive and requires a large amount of GPU memory for LLMs. To improve the computational efficiency, LoRA decomposes $\Delta{W}$ into two small matrices $B \in \mathbb{R}^{d \times r}$ and $A \in \mathbb {R }^{r\times k}$, i.e.,
\begin{equation}
W = W_0 + \alpha BA
\end{equation}
where $r \ll min\{d, k\}$, $B$ and $A$ are initialized with a random Gaussian distribution and zero respectively, $\alpha$ represents the scaling factor that controls the strength of updates.
The parameter number of LoRA is $r \times (d+k)$, which is significantly less than $d \times k$.  Figure~\ref{fig:lora} (a) and (b) compare the structures of full fine-tuning and LoRA.

LoRA is highly \textbf{parameter efficient} for it updates only a small subset of model parameters, which reduces the memory and computational requirements for fine-tuning without increasing inference latency~\cite{fomenko2024note}. Furthermore, The parameter efficiency can be further improved by extending from the low-rank matrix to low-rank tensor~\cite{journals/corr/abs-2402-01376} or combining with the Kronecker decomposition~\cite{journals/corr/abs-2212-10650, conf/aaai/HeLZYW23}. Except for parameter efficiency, LoRA is also \textbf{pluggable} for the LoRA parameters that can be separated from the model after training. The pluggable character of LoRA enables it to be shared and reused by multiple users~\cite{s-lora}. When we have LoRA modules for multiple tasks, we can combine these modules and expect a proper \textbf{cross-task generalization} performance~\cite{lorahub}. Besides, the low-rank mechanism of LoRA is \textbf{compatible} with other parameter-efficient methods, such as adapter ~\cite{conf/nips/MahabadiHR21,conf/acl/LiaoMM23}. 
Besides, LoRA can achieve \textbf{proper downstream adaptation performance} on various downstream tasks. For example, on MMLU~\cite{hendrycks2020measuring} benchmark, comparing with full fine-tuning, fine-tuning with LoRA can achieve comparable or even better performance across 57 tasks~\cite{journals/corr/abs-2403-17919}. 

In practice, for a Transformer-based LLM, the dense layers typically consist of two types of weight matrices: the projection matrices in attention modules and feed-forward neural~(FFN) modules. The experiments mentioned above are conducted based on the original LoRA settings, applying it to the query and value weight matrices in the attention modules. It is worth mentioning that subsequent work shows that applying it to the FFN layers can further improve model performance~\cite{conf/iclr/HeZMBN22}.

\subsection{Theoretical Analysis}
\label{subsec:theoretical_analysis}

To understand why LoRA is effective and how LoRA can be more effective, several works have provided theoretical analyses from various aspects. To answer the question that why LoRA is effective, Malladi et al.~\cite{conf/icml/MalladiWYCA23} analyze the fine-tuning dynamics of LoRA from the kernel view and demonstrate that in the lazy regime, LoRA fine-tuning is nearly equivalent to full fine-tuning. 
Besides, Zeng et al.~\cite{journals/corr/abs-2310-17513} provides a theoretical analysis of the LoRA's expressive power for both fully connected neural networks~(FNNs) and Transformer networks~(TFNs).
They proved that LoRA can adapt any model $f$ to accurately represent any smaller target model $\bar f$ if LoRA-rank $\geq$ (width of $f$) $\times$ $\frac{depth\, of \, \bar f}{depth\, of\, f}$ under a mild assumption, where the depth and width are the number of layers and the number of neurons of the layer having the largest number of neurons, respectively. Moreover, they quantify the approximation error when the LoRA-rank falls below this threshold.
Regarding TFNs, they showed that any model can be adapted to a target model of equivalent size using a rank-$\left(\frac{\text{embedding size}}{2}\right)$ for LoRA.
Additionally, Koubbi et al.~\cite{journals/corr/abs-2402-15415} utilize the mathematical framework for Transformers established by ~\cite{journals/corr/abs-2312-10794,conf/nips/GeshkovskiLPR23,conf/aistats/SanderABP22} to investigate how variations in attention parameters and initial token values impact the structural dynamics of token clusters.
% low-rank perturbations in attention parameters affect. 

As to the question that how LoRA can be more effective, Jang et al.~\cite{journals/corr/abs-2402-11867} analyze the fine-tuning of LoRA within the neural tangent kernel (NTK)~\cite{conf/nips/JacotHG18} framework when $N$ data points are available. They demonstrate that employing a rank $r \gtrsim \sqrt{N}$ in LoRA helps to avoid spurious local minima and facilitates the discovery of low-rank solutions that exhibit good generalization.
Besides, Zhu et al.~\cite{journals/corr/abs-2402-16842} observe that the project-down matrix $A$ is utilized for extracting features from the input, while the project-up matrix $B$ employs these features to create the desired output. Based on this observation, they 
demonstrate that freezing the project-down matrix $A$ while tuning only the project-up matrix $B$ leads to better generalization compared to tuning both matrices, in addition to achieving a $2 \times$ reduction in parameters.

\subsection{Efficiency in Practice} 

The computational efficiency of LoRA is significantly higher than that for full fine-tuning. Taking fine-tuning the dense weight matrix of the first FFN layer in LLaMA2-7B as an example, full fine-tuning needs to fine-tune  $11,008 \times 4,096 = 45,088,768$ parameters while LoRA only needs to tune $(11,008 \times 4) + (4 \times 4,096) = 60,416$ parameters when $r=4$. For this layer, LoRA only adjusts nearly one-thousandth of the parameters compared to full fine-tuning.
% % 
% Taking fine-tuning the first FFN layer of LLaMA2-7B as an example, we have the dense weight matrix $W_0$ of the FFN, which has $11,008 \times 4,096 = 45,088,768$ parameters.
% % 
% While the LoRA layer, we set LoRA rank $r$ to 4, $A$ and $B$ together have $(11,008 \times 4) + (4 \times 4,096) = 60,416$ parameters. 
% % 
% Therefore, for the dense layer, we reduce the number of trainable parameters from $45,088,768$ to only $60,416$. It can be seen that LoRA has a very high parameter efficiency. 

% We can significantly reduce the number of training parameters by selecting a lower rank $r$ (usually $r \in \{2,4,8,16,32\}$). For example, if $W$ is a $d \times d$ matrix, fine-tuning all parameters in $W$ would involve $d^2$ parameter updates. However, when the sizes of $B$ and $A$ are respectively $d \times r$ and $r \times d$, the total number of parameters decreases to $2dr$, where $r << d$. 

% Here is a specific example showing the situation of parameter quantity when fine-tuning a single weight matrix:
% 
% Assuming $n$ is $768$, and $r$ is 4. $W_0$ has $768 \times 768 = 589,824$ parameters, while the LoRA layer, $A$, and $B$ together have $(768 \times 4) + (4 \times 768) = 6,144$ parameters. Therefore, for the dense layer, we reduce from having $589,824$ trainable parameters to having only 6,144 trainable parameters.
% 
% Of course, subsequent work~\cite{conf/iclr/HeZMBN22} found that applying LoRA to more different matrices (such as the projection matrix of FFN) can further improve model performance.

LoRA can significantly decrease the memory usage of fine-tuning an LLM, which can be divided into four parts: (1) Model Memory: the memory required to store the model weights; (2) Activation Memory: the memory occupied by intermediate activations during forward propagation. It mainly depends on factors such as batch size and sequence length; (3) Gradient Memory: the memory required to store gradients during backpropagation. The gradients are only calculated for trainable parameters; (4) Optimization Memory: the memory used to store optimizer states. For example, the Adam optimizer stores the ``first moment'' and ``second moment'' of trainable parameters.

Pan et al.~\cite{journals/corr/abs-2403-17919} provides a comprehensive empirical comparison between full fine-tuning and LoRA fine-tuning on an LLaMA2-7B model with batch size 1, utilizing a single NVIDIA RTX4090 (24GB) GPU.
According to this study, full fine-tuning requires approximately 60GB of memory, which exceeds the capacity of an RTX4090 GPU;
by contrast, LoRA fine-tuning only needs about 23GB of memory. LoRA significantly reduces memory usage and makes fine-tuning LLaMA2-7B feasible on a single NVIDIA RTX4090 (24GB) GPU. 
Specifically, due to fewer trainable parameters, both optimization memory and gradient memory decrease significantly by approximately 25GB and 14GB respectively. On the other hand, while LoRA introduces additional ``incremental parameters'' resulting in slight increases in activation memory and weight memory (totaling about 2GB), this increase is negligible when considering the overall reduction in memory. Moreover, reducing memory brings an acceleration of forward propagation. LoRA is $1.9 \times$ times faster compared to full fine-tuning.

\subsection{Beyond Fine-tuning}
\label{subsec:beyond_ft}

Besides fine-tuning, LoRA can be applied to other learning paradigms, such as pre-training~\cite{lialin2023relora, journals/corr/abs-2402-16828} and continual training~\cite{journals/corr/abs-2404-00228}. For pre-training,  \textbf{ReLoRA}~\cite{lialin2023relora} and \textbf{MoRA}~\cite{jiang2024mora} are proposed to use low-rank updates to train high-rank networks; moreover, \textbf{LTE}~\cite{journals/corr/abs-2402-16828} is proposed to perform parallel training of multiple low-rank heads across computing nodes to minimize the need for frequent synchronization, which facilitates the utilization of LoRA in pre-training. As for continual training, there are several methods have been proposed to address the catastrophic forgetting problem. \textbf{InfLoRA}~\cite{journals/corr/abs-2404-00228} addresses catastrophic forgetting by reparameterizing pre-trained weights with a minimal set of parameters in a subspace. \textbf{GS-LoRA}~\cite{journals/corr/abs-2403-11530} uses group sparse regularization to automatically select specific LoRA groups while zeroing out others to mitigate catastrophic forgetting effects. \textbf{I-LoRA}~\cite{journals/corr/abs-2402-18865} leverages dual-memory experience replay combined with LoRA parameter interpolation to combat catastrophic forgetting.

Furthermore, LoRA can be used to overcome the limited context size for LLMs ~\cite{journals/corr/abs-2309-12307,journals/corr/zhang2024sinklora}. For instance,
\textbf{LongLoRA}~\cite{journals/corr/abs-2309-12307} successfully computaitional efficiently extends the context window of LLaMA2-7B~\cite{journals/corr/abs-2307-09288} from 4k to 100k tokens by combining LoRA with shifted sparse attention.
However, LongLoRA does not match the efficiency of vanilla attention due to chaotic attention head structures and unnecessary information exchange between token groups. 
To address these issues, \textbf{SinkLoRA}~\cite{journals/corr/zhang2024sinklora} introduces Sink Fixed Attention~(SF-Attn) to proportionally returns cyclically shifted groups of attention heads to their un-shifted state and achieves proper performance.

\section{Downstream Adaptation Improving}
\label{sec:Downstream_Adaptation}
Although LoRA can achieve proper adaptation performance on some downstream tasks, there is still a performance gap between LoRA and full fine-tuning on many downstream tasks, such as mathematical reasoning~\cite{journals/natmi/Ding23,journals/corr/abs-2403-03507,journals/corr/abs-2405-09673}. To fill this gap, many methods are proposed to further improve the downstream adaptation performance of LoRA. Typically, existing methods improve the downstream adaptation performance from the following perspectives: (1) breaking the low-rank bottleneck, refer to Figure~\ref{fig:lora} (c); (2) adaptively allocating the ranks of different LoRA modules, refer to Figure~\ref{fig:lora} (d); (3) optimizing the learning procedure of LoRA; (4) combining with other learning paradigms. In this section, we introduce these four types of methods respectively.
% % 
% While LoRA can effectively lower the cost of fine-tuning models and show versatility across various domains and tasks, it does have some drawbacks. 
% % 
% For instance, it is challenging to determine an approximate rank for LoRA, and its low-rank nature restricts its performance to be on par with full fine-tuning.
% % 
% Furthermore, the original LoRA's training dynamics can be enhanced to boost training stability, reduce overfitting, and even allow for optimization without the need for gradients.
% % 
% In this section, we introduce the different variants of LoRA, which can be divided into three main aspects: structural refinement and low-rank dilemma and weight decomposition method.
% % 
% Next, we will present the details of each improvement strategy.

\subsection{Breaking the Low-rank Bottleneck}
\label{subsec:Breaking_Bottleneck}

The low-rank updates enable LoRA to be parameter efficient; however, it restricts LLMs' ability to memorize downstream knowledge and generalization on downstream tasks ~\cite{jiang2024mora,han2024sltrain,journals/corr/abs-2401-10341,journals/corr/abs-2403-03507,journals/corr/abs-2405-09673}. This low-rank limitation causes inferior performance of LoRA in knowledge- and skill-intensive domains comparing to full-fine tuning, such as code and math. Experimental study \cite{journals/corr/abs-2405-09673} demonstrates that the rank for full fine-tuning is significant ~(10-100 $\times$) higher than that for LoRA, and increasing the rank of LoRA updation can narrow the performance gap between LoRA and full fine-tuning. To increase the rank of LoRA and improve its performance, several methods have been proposed~\cite{lialin2023relora,journals/corr/abs-2401-04151,journals/corr/abs-2309-02411,journals/corr/abs-2402-16141}, which typically increase the rank through (1) stacking LoRAs along learning iterations; (2) updating as gradient compressors; (3) co-updating LLM and LoRA modules during fine-tuning.

\subsubsection{Stacking LoRAs along Fine-tuning}
\label{subsubsec:Stacking_Along_Ft}

Matrix rank is subadditive, i.e., $rank(M_1 + M_2) \leq rank(M_1) + rank(M_2)$ for matrices $M_1$ and $M_2$ that have the same size. Based on the subadditivity, we can aggregate multiple LoRA modules together to increase the rank and break the low-rank bottleneck. Following this idea, \textbf{ReLoRA}~\cite{lialin2023relora} proposes a merge-and-reinit procedure for LoRA, which periodically merges the LoRA modules to the LLM and then reinitializes the LoRA modules during fine-tuning. It equals stacking multiple LoRA modules along with fine-tuning and can increase the rank of the overall updates. Similarly, \textbf{COLA}~\cite{journals/corr/abs-2401-04151} proposes another merge-and-reinit method based on Frank-Wolfe algorithm~\cite{frank1956algorithm}. However, \textbf{MELoRA}~\cite{journals/corr/abs-2402-17263} points out that the merge-and-reinit procedure does not necessarily guarantee an increase in rank, because there can be overlap between the series of LoRA modules along fine-tuning. To solve this problem, MELoRA proposes to decompose the LoRA modules into smaller mini LoRAs and then parallelly stack these mini LoRAs, whose effectiveness in increasing the rank is theoretically verified.  

\subsubsection{Updating as Gradient Compressor}
\label{subsubsec:Avoiding_Gradient_Compression}

The above methods break the low-rank bottleneck in the parameter space. As a supplement, \textbf{FLoRA}~\cite{journals/corr/abs-2402-03293} finds that LoRA performs a fixed random projection to compress gradients and restricts the total weight matrix change to low-rank.
To overcome this low-rank bottleneck in gradient space, \textbf{FLoRA} proposes to resample the random projection, which is demonstrated to largely recover the performance of full-matrix SGD.
% 
% \textbf{FLoRA}~\cite{journals/corr/abs-2402-03293} finds that LoRA functions as a form of gradient compression via random projections. 
% % 
% By resampling the projection matrices, FLORA achieves high-rank updates while maintaining sublinear space complexity for the optimization states.

\subsubsection{Co-updating LLM and LoRA}
\label{subsubsec:Co-updating_LLM_and_LoRA}

The above two kinds of methods focus on improving the representation ability of LoRA itself. Different from them, \textbf{Delta-LoRA}~\cite{journals/corr/abs-2309-02411} proposes to jointly update the LLM and LoRA modules, which directly updates the high-rank LLM and can gain better representation capable than updating LoRA independently. It updates the LLM based on the difference between two LoRA modules of two consecutive iterations, which enables it to update the LLM without any extra memory.

\subsection{Dynamic Rank Allocation}
\label{subsec:dynamic_rank_allocation}

For the rank of LoRA, higher is not always better. The abundant LoRA ranks may cause degeneration in both performance and efficiency. 
Furthermore, the importance of weights can vary across different layers of a Transformer model during fine-tuning, requiring different ranks for each layer.~\cite{conf/iclr/ZhangCBH0CZ23,conf/emnlp/DingLWCZL023,mao2024dora,valipour2022dylora}. 
Therefore, assigning the same rank to LoRA modules of different layers is not the optimal choice. It is better to adaptively allocate ranks to LoRA modules of different layers. Existing methods adaptively allocate ranks for LoRA modules from the perspectives of (1) singular value decomposition (SVD); (2) single-rank decomposition (SRD); (3) rank sampling.

\subsubsection{SVD-based Methods}
\label{subsubsec:svd_based_methods}

Decomposing a matrix with singular value decomposition (SVD) and selectively truncating its singular values is an effective way to control the rank of the matrix. Inspire by SVD, we can decompose the LoRA parameter matrix $BA$ into an SVD form, i.e, $P\Lambda Q$ where $P$ and $Q$ are orthogonal and $\Lambda$ is a non-negative diagonal matrix. 
By controlling the elements in $\Lambda$, we can control the rank of $BA$ and allocate ranks for LoRA modules. Following this idea, several rank allocation methods approximate the SVD decomposition for $BA$ and allocate the ranks by filtering the diagonal matrix. 
For instance, \textbf{AdaLoRA}~\cite{conf/iclr/ZhangCBH0CZ23} approximates the SVD decomposition by regularizing the orthogonality of $P$ and $Q$. Then, it drops unimportant singular values based on novel importance scoring methods. Similarly, \textbf{SaLoRA}~\cite{Hu2023Structure} also introduces an orthogonality regularization for $P$ and $Q$; by contrast, it drops unimportant singular values based on the $L_0$ norm. However, the above methods are not efficient enough for they 
start with a high rank and then reduce the rank iteratively, which brings a pre-defined budget ~\cite{journals/corr/abs-2308-12043}. To solve this problem, \textbf{IncreLoRA}~\cite{journals/corr/abs-2308-12043} proposes to start from a single rank and then automatically increase the rank based on a heuristic importance score, where the orthogonality regularization is also involved while the elements in $\Lambda$ is not required to be non-negative.

\subsubsection{SRD-based Methods}
\label{subsubsec:srd_based_methods}

However, the orthogonality regularization brings unignorable computational costs for LoRA and degenerates its efficiency. To address this problem, several methods omit the orthogonality requirement of SVD and directly decompose $BA$ into single-rank components. Then, they allocate the ranks by selecting the proper components. \textbf{DoRA~(Dynamic Low-Rank Adaptation)}~\cite{mao2024dora}  proposes to decompose the LoRA parameter matrix $BA$ into single-rank components and prunes the components based on a heuristic importance score. Similarly,  \textbf{AutoLoRA}~\cite{journals/corr/abs-2403-09113} also decomposes the LoRA parameter matrix $BA$ into single-rank components, but it prunes the components based on meta-learning.
\textbf{SoRA}~\cite{conf/emnlp/DingLWCZL023} eliminates the orthogonality regularization and filters columns and rows of $P$ and $Q$ (their combination can be regarded as single-rank components) by directly controlling the diagonal matrix. It controls the diagonal matrix by formulating them as a set of learnable gating units which are updated in the fine-tuning procedure. 
\textbf{ALoRA}~\cite{journals/corr/abs-2403-16187} also filters the components by using gating units; by contrast, it learns the gating units based on neural architecture search~\cite{journals/jmlr/ElskenMH19}.

\subsubsection{Rank Sampling-based Methods}
\label{subsubsec:rank_sampling_based_methods}

In the SVD parameterization- and component-wise decomposition-based methods, we need to spend the extra computational costs to search proper ranks. To avoid the extra cost, \textbf{DyLoRA}~\cite{valipour2022dylora} points out that we can allocate ranks directly by random sampling. In each training step, it samples a value $b$ from a pre-defined discrete distribution and allocates $b$ as the rank. Then, the matrices \( A \) and \( B \) are truncated to rank-$b$. In the fine-tuning procedure, only the parameters on the $b$-th row of \( A \) and $b$-th column of \( B \) are tunable while other parameters are frozen. Besides, the distribution can be defined based on users' preferences. 

% The rank sampling method assumes that the LoRA rank is not a single fixed value, but rather sampled from a probability distribution. For instance, 
% % 
% \textbf{DyLoRA}~\cite{conf/eacl/ValipourRKG23} trains the LoRA modules on a range of rank choices from a pre-defined discrete distribution $P_B(\cdot)$, without changing the training budget.
% % 
% In particular, DyLoRA dynamically samples $r$ from the distribution $P_B(\cdot)$, where $r$ ranges from $r_{min}$ to $r_{max}$, at every training iteration.
% % 
% Matrices \( A \) and \( B \) are adjusted to the chosen rank \( r \), yielding truncated versions: \( A_{\downarrow r} = A[:r, :] \) and \( B_{\downarrow r} = B[:, :r] \).
% % 
% During the iteration, forward and backward propagation will be constrained by $A_{\downarrow r}$ and $B_{\downarrow r}$ instead of by the original matrices $A$ and $B$.
% % 
% Without multiple training and optimal rank searching, DyLoRA can support a wider range of ranks for the specific task.

\subsection{Optimizing the Learning Procedure}
\label{subsec:opt_learning_procedure}

In practice, LoRA converges more slowly than full fine-tuning. Moreover, it is also sensitive to hyperparameters and suffers from overfitting. These issues affect LoRA's efficiency and hinder its downstream adaptation performance. To address these issues, researchers have developed several approaches to optimize the learning procedure of LoRA, which can be categorized into the following three types: (1) Initialization Improvement; (2) Gradient Update Optimization; (3) Overfitting Mitigation.

\subsubsection{Initialization Improvement}
\label{subsubsec:init_improvement}

% To start from the pretrained model as initialization for finetuning, LoRA usually initializes matrices A and B using Gaussian noise and zeros.
LoRA usually initializes its parameter matrices A and B using Gaussian noise and zeros respectively.
There are two simple schemes: Init[A], which sets matrix B to zero and randomly initializes matrix A, and Init[B], which does the reverse. Literature ~\cite{hayou2024impact} compares these two schemes and concludes that Init[A] is better through theoretical analysis. It reveals that Init[A] allows using a larger learning rate without causing instability, making the learning process more efficient. However, even with init[A], this random initialization method still results in small initial gradients, leading to slower convergence. To solve this, \textbf{PiSSA}~\cite{PiSSA} initializes LoRA with the principal singular components of the pre-trained matrix. Since principal singular components represent the most significant directions in the matrix, aligning the initial weights with these components can accelerate convergence and improve performance. In contrast, \textbf{MiLoRA}~\cite{MiLoRA} initializes LoRA with the minor singular components. Given that random initialization of low-rank matrices can interfere with the important features learned in the pre-trained matrix, it reduces this interference to improve overall performance while adapting to new tasks.

\subsubsection{Gradient Update Optimization}
\label{subsubsec:gradient_update_opt}

% To further enhance the convergence and reliability of LoRA, several studies have proposed improvements from the perspective of gradient updates. 
To further enhance the convergence and reliability of LoRA, several studies have proposed improvements from the perspective of gradient updates. 
~\cite{zhang2024riemannian} introduces a scaled gradient method based on Riemannian optimization, which incorporates an $r \times r$ preconditioner item in the gradient update step to improve the convergence and hyperparameter robustness of LoRA. Through theoretical analysis, \textbf{LoRA+}~\cite{lora+} discovered the necessity of setting a proportional learning rate for matrices A and B to achieve stable feature learning and accelerate convergence. \textbf{ResLoRA}~\cite{ResLoRA} introduced residual connections into LoRA to optimize the gradient propagation path, speeding up training convergence and enhancing model performance. Similarly, \textbf{SIBO}~\cite{SIBO} mitigate over-smoothing by injecting residual connections of initial token representations into LoRA's input. Additionally, to further reduce computational resources, literature ~\cite{derivative-free} employs gradient-free optimization methods such as CMA-ES and FWA to optimize LoRA, demonstrating competitive performance in few-shot NLU tasks. Besides, \textbf{DoRA}~(Weight-Decomposed Low-Rank Adaptation)~\cite{DoRA} constrains the gradient update, focusing on the directional change of the parameter. It decomposes pre-trained weight into two components, direction and magnitude, and applies LoRA only to the direction component to enhance training stability.

\subsubsection{Overfitting Mitigation}
\label{subsubsec:overfitting_mitigation}

Although LoRA effectively reduces the number of trainable parameters compared to full fine-tuning, some studies have shown that LoRA is also prone to overfitting~\cite{HiddenKey}, which contradicts previous views. To address this issue, \textbf{BiLoRA}~\cite{BiLoRA} adopts a bi-level optimization strategy. It alternately trains the singular vectors and singular values of the low-rank increment matrix on different subsets of the training data. This approach avoids the simultaneous optimization of parameters at different levels on a single dataset, thus mitigating overfitting. In addition, literature ~\cite{lora-dropout} applies dropout to LoRA parameters to reduce overfitting, while \textbf{HiddenKey}~\cite{HiddenKey} employs column-wise dropout for attention layers and element-wise dropout for feedforward layers.

\subsection{Combining with other Learning Paradigms}
\label{subsec:combining_with_other_learning_paradigms}

LoRA is compatible with other learning paradigms, such as Bayesian Learning, In-context Learning and Active Learning. Combining LoRA with these learning paradigms can address several problems that hurt the downstream adaptation performance. For example, combining with Bayesian Learning, \textbf{Laplace-LoRA}~\cite{journals/corr/abs-2308-13111} can relieve the overconfidence phenomenon that happened in downstream adaptation. Combining with In-context Learning,  \textbf{PILLOW}~\cite{conf/emnlp/QiTSQXQ23} aims to solve the low-resource dilemmas existing in some downstream tasks. Combining with Active Learning, \textbf{STAR}~\cite{journals/corr/abs-2403-01165} can effectively improve the data efficiency. 
\newline

At last, to illustrate the performance difference between LoRA and some of its variants, we report their performance for RoBERTa-base~\cite{liu2019roberta} model on the GLUE benchmark~\cite{wang2018glue} in Table~\ref{tab:lora-performance}. These results are derived from previous studies~\cite{conf/iclr/HuSWALWWC22, gao2024parameter, journals/corr/abs-2403-09113, journals/corr/abs-2310-17513, BiLoRA}.

\begin{table*}
\centering
\caption{Performance of LoRA and its variants for RoBERTa-base model on the GLUE benchmark. We report Matthew's correlation for CoLA, Pearson correlation for STS-B, and accuracy for the other datasets. The results are reported according to the results reported in literature \cite{conf/iclr/HuSWALWWC22,gao2024parameter,journals/corr/abs-2403-09113,BiLoRA,vb-lora}.}
\label{tab:lora-performance}
\vspace{0.25cm}
% \resizebox{\textwidth}{15mm}{
\setlength{\tabcolsep}{4.8mm}{
% \fontsize{2pt}{30pt}\selectfont
\begin{tabular}{l|r|cccccc}
\toprule
\textbf{Method}                                                               & \textbf{\# Params} & \textbf{SST-2} & \textbf{MPRC} & \textbf{CoLA} & \textbf{QNLI} & \textbf{RTE} & \textbf{STS-B} \\ \midrule
Tied-LoRA~\cite{renduchintala2024tied}       & 0.043M             & 94.4           & 88.5          & 61.9          & 92.0            & 76.2         & 89.8          \\
AutoLoRA~\cite{journals/corr/abs-2403-09113} & 0.3M               & 94.9           & 89.4          & 61.3          & 92.9          & 77.0           & 90.8          \\
DyLoRA~\cite{valipour2022dylora}        & 0.3M               & 94.3           & 89.5         & 61.1         & 92.2         & 78.7         & 91.1         \\
AdaLoRA~\cite{conf/iclr/ZhangCBH0CZ23}       & 0.3M               & 94.5           & 88.7          & 62.0            & 93.1          & 81.0           & 90.5           \\
FourierFT~\cite{gao2024parameter}            & 0.024M             & 94.2           & 90.0            & 63.8          & 92.2          & 79.1         & 90.8        \\
VeRA~\cite{vera}                             & 0.043M             & 94.6           & 89.5          & \textbf{65.6}          & 91.8          & 78.7         & 90.7        \\
Full Fine-tuning~\cite{conf/iclr/HuSWALWWC22}              & 125M               & 94.8           & 90.2          & 63.6          & 92.8          & 78.7         & 91.2    \\
LoRA~\cite{conf/iclr/HuSWALWWC22}            & 0.3M             & \textbf{95.1}           & 89.7          & 63.4          & \textbf{93.3}          & 78.4         & 91.5    \\
VB-LoRA~\cite{vb-lora}                       & 0.023M             & 94.4           & 89.5          & 63.3          & 92.2          & 82.3         & 90.8           \\
BiLoRA~\cite{BiLoRA}                         & 0.3M               & \textbf{95.1}           & \textbf{91.7}          & 64.8          & \textbf{93.3}          & \textbf{87.2}         & \textbf{91.7}         \\
\bottomrule
\end{tabular}}
\end{table*}

\section{Cross-task Generalization}
\label{sec:Cross-task_Generalization}

\begin{figure*}[ht]
    \centering
    \includegraphics[width=1.0\linewidth]{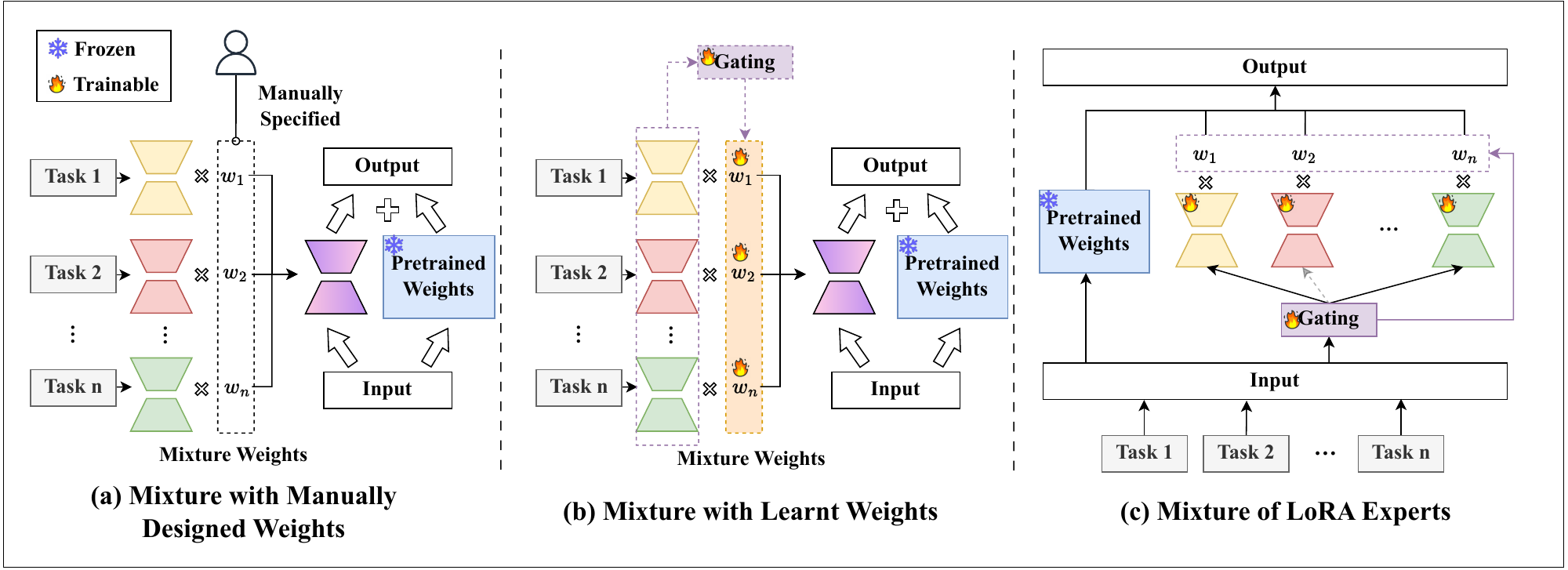}
    \caption{An illustration of LoRA mixture methods.}
    \label{fig:lora_mixture}
\end{figure*}

LoRA's pluggable nature enables users to accumulate LoRA plugins for different tasks. For example, on Hugging Face platform, there are more than 20,000 LoRA plugins compatible with various LLMs for different tasks. These accumulated LoRA plugins can not only be utilized independently but also be mixed to achieve cross-task generalization\cite{lorahub}. Mixing multiple LoRA plugins together, namely LoRA mixture, has been widely applied in areas requiring cross-task generalization, such as multi-task learning, domain adaptation, and continual learning. Existing LoRA mixture methods can be categorized into (1) mixture with manually designed weights; (2) mixture with learnt weights; (3) mixture of LoRA experts. This section introduces each category of methods respectively, as shown in Fig.~\ref{fig:lora_mixture}.

\subsection{Mixture with Manually Designed Weights}
\label{subsec:Mixture with Manually Designed Weights}

Early LoRA mixture methods attempt to linearly combine different LoRA modules with manually designed weights. Some research demonstrates that we can achieve proper cross-task generalization ability by simply averaging LoRA modules or their related outputs \cite{lora-ensembles,lora-retriever,construct-vl}. Furthermore, several methods have been proposed to further improve the performance of the LoRA mixture via adopting manually designed weights. For example, \textbf{ControlPE}~\cite{sun2023or}, ~\cite{composing} and \cite{task-arithmetic} set the weight factors as hyperparameters, and ControlPE uses hyperparameter search to determine the optimal combination of two LoRA modules. Additionally, \textbf{Token-level Adaptation}~\cite{token-level} utilizes cosine similarity between the input feature and the adapter dataset center as weight factors, while \textbf{BYOM}~\cite{byom} applies basic model fusion methods such as Task Arithmetic, Fisher-Merging, and RegMean.

Mixture with manually designed weights can quickly mix multiple LoRAs without extra training, which demonstrates simplicity and computational efficiency. However, it often fails to find the optimal weights, leading to unstable performance and limited generalization. Subsequently, researchers have explored using learning-based methods to achieve more precise and adaptive mixtures.

\subsection{Mixture with Learnt Weights}
\label{subsec:Mixture with Learnt Weights}

To learn the optimal mixture weights, several methods have been proposed at task level, instance level and token level to meet different needs.
Task-level methods focus on enhancing task transferability, which can be either gradient-based, such as~\cite{does}, or gradient-free, as seen in \textbf{LoRAHub}~\cite{lorahub}. LoRAHub employs a black-box algorithm named CMA-ES~\cite{cma-es} to optimize weight factors for LoRA modules, simplifying the training process. Later, \textbf{ComPEFT}~\cite{compeft} and \textbf{L-LoRA}~\cite{l-lora} use LoRAHub to mix quantized LoRA modules, further improving computational efficiency.

Compared to task-level methods, instance-level and token-level methods can provide flexibility and precision for complex inputs. For multimodal instruction tuning, \textbf{MixLoRA}~\cite{multimodal} dynamically chooses appropriate low-rank decomposition vectors based on the input instance, which are then integrated into LoRA matrices for training. To conduct protein mechanics analysis and design tasks, \textbf{X-LoRA}~\cite{x-lora} develops a dynamic gating mechanism to assign weights for LoRA modules at the token level and layer granularity. These approaches demonstrate better performance in specific tasks or application scenarios.

\subsection{Mixture of LoRA Experts}
\label{subsec:Mixture of LoRA Experts}

When the LoRA modules are trainable, we can jointly learn the mixture weights and the LoRA modules, which can further improve the performance of the LoRA mixture. To jointly learn the mixture weights and LoRA modules, Mixture of LoRA Experts (LoRA MoE) is a natural choice, where each LoRA module acts as an expert, while a router network typically assigns the mixture weights. LoRA MoE has been proven to be effective in many tasks, such as continual learning\cite{moral, loramoe}, vision-language tasks\cite{mocle} and multi-task medical applications\cite{moelora}.

Existing methods improve the performance of LoRA MoE from the perspectives of initialization, task relationship management and efficiency. 
For initialization, \textbf{Mixture-of-LoRAs}~\cite{mixture-of-loras} first trains multiple LoRAs separately as initialization and then optimizes the router and LoRAs jointly. \textbf{MultiLoRA}~\cite{multilora} proposes refining the initialization to reduce parameter dependency, which can yield more balanced unitary subspaces. As for task balance,  \textbf{MLoRE}~\cite{multi-task} adds a low-rank convolution path in the MoE structure to capture global task relationships. 
\textbf{MTLoRA}~\cite{mtlora} adopts both task-agnostic and task-specific LoRA modules to address task conflicts.
For efficiency, \textbf{MoLA}~\cite{higher} adaptively allocates different numbers of LoRA experts to different layers of the Transformer model to save the number of LoRA modules. \textbf{LLaVA-MoLE}~\cite{llava-mole} and \textbf{SiRA}~\cite{sira} leverage sparse computation to reduce computational cost. Additionally, \textbf{Octavius}\cite{octavius} 
sparsely activates independent LoRA experts with instance-level instructions to mitigate task interference and improve efficiency. \textbf{Fast LoRA}\cite{FLoRA} allows each sample in a minibatch to have its unique low-rank adapters, enabling efficient batching.
 
Besides, some methods are not explicitly based on MoE but follow MoE ideas. For example, \textbf{I-LoRA}~\cite{journals/corr/abs-2402-18865} uses two LoRAs to manage long-term and short-term memory for continual learning, respectively.

\section{Efficiency Improving}
\label{sec:Computational_Efficiency}
% LoRA computational efficiency optimization refers to improving the computation process and hardware resource utilization to enhance the speed of model training and inference. In this section, we will introduce three computational efficiency optimization methods: quantization, parameter-fixed optimization, and parallel and distributed strategies.
% Although LoRA has already shown significant effects in accelerating model training and inference, there is still room for further optimization. To maximize computational and memory efficiency, various methods can be adopted, such as quantization, parameter-reduced optimization, and LoRA serving system optimization. The following sections will detail these existing methods.

\begin{figure*}
    \centering
    \includegraphics[width=1.0\linewidth]{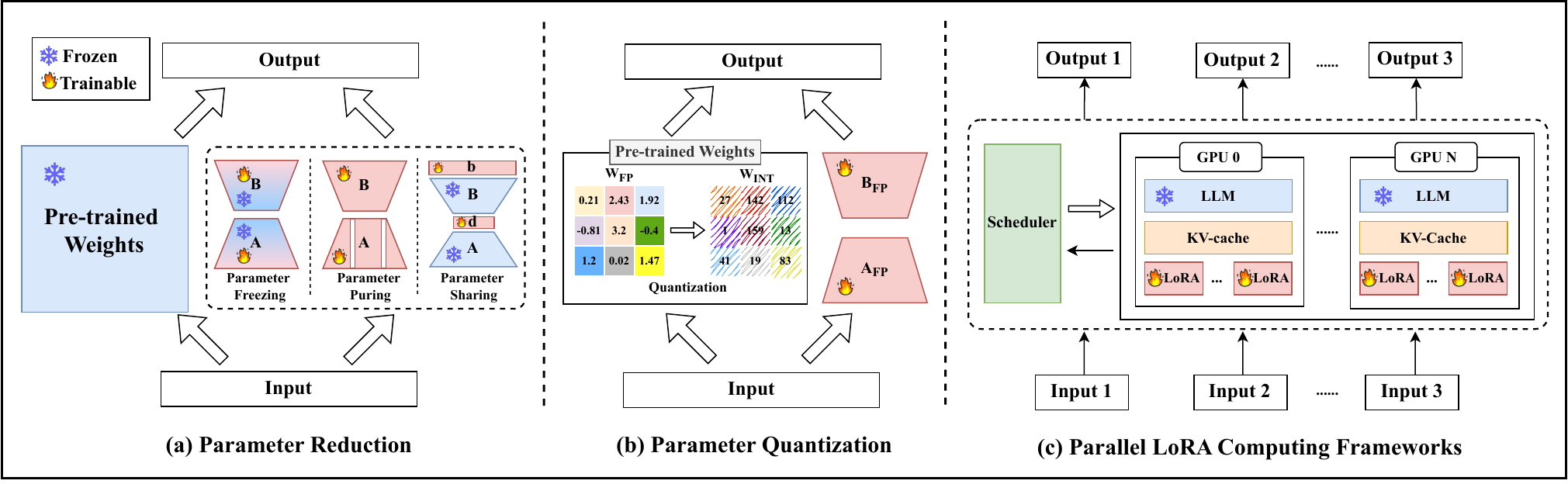}
    \caption{An illustration of efficiency improving methods.}
    \label{fig:lora_computing}
\end{figure*}

With the popularization of LLMs, the demand for training and running LoRA modules increases rapidly. This increasing demand brings an unignorable computational burden; thus, for LoRA, the smaller, the faster, the better. To meet this demand, existing methods improve the computational efficiency of LoRA from the perspectives of (1) parameter reduction; (2) parameter quantization; (3) parallel LoRA computing frameworks. This section introduces each category of methods, as illustrated in Fig.~\ref{fig:lora_computing}.

\subsection{Parameter Reduction}
\label{subsubsec:parameter-reduction}

LoRA significantly reduces the number of tunable parameters for fine-tuning LLMs. However, it still requires expensive activation memory to update low-rank matrices. To further reduce the memory cost, existing methods reduce the number of tunable parameters of LoRA via parameter freezing, parameter pruning, and parameter sharing.

% LoRA significantly reduces the number of tunable parameters for fine-tuning LLMs. However, it still requires expensive activation memory to update low-rank weights. To optimize further, researchers reduce the number of trainable parameters in LoRA. This lowers memory and computational costs while maintaining model performance as much as possible.Existing parameter-reduced methods can be categorized into the following types:(1)Partial Parameter Freezing, (2)Pruning, (3)SVD-Based Methods, (4)Gradient Compression, and (5)Parameter Sharing.

\subsubsection{Parameter Freezing}
\label{subsubsec:parameter-freezing}

Parameter freezing methods reduce the number of tunable parameters for LoRA via freezing some of its parameters. They can be divided into two categories: intra-parameter methods and extra-parameter methods.

The intra-parameter methods tune a subset of parameters of LoRA while freezing the others. \textbf{LoRA-SP}\cite{lora-sp} randomly selects half of the LoRA parameters to freeze during fine-tuning. \textbf{LoRA-FA}\cite{lora-fa}freezes the down-projection weights and updates the up-projection weights in each layer of LoRA. \textbf{AFLoRA}\cite{aflora} constructs a low-rank trainable path and gradually freezes parameters during training LoRA. Additionally, \textbf{DropBP}\cite{dropbp} accelerates the training process by randomly dropping some LoRA gradient calculations during backpropagation.

By contrast, the extra-parameter methods introduce and tune a set of extra parameters while freezing the original parameters of LoRA. Most of them are proposed based on Singular Value Decomposition(SVD). \textbf{LoRA-XS}\cite{lora-xs} adds a small \(r \times r\) weight matrix between frozen LoRA matrices, which are constructed using the SVD of the original weight matrix; then it tunes only the \(r \times r\) weight matrices in fine-tuning. Similarly, \textbf{BYOM-LoRA}\cite{byom} adopts SVD to compress LoRA matrices for multi-task models.

\subsubsection{Parameter Pruning}
\label{subsubsec:parameter-pruning}

Parameter pruning methods aim to remove unimportant LoRA parameters during training and inference. They prune parameters by either pruning LoRA independently or jointly pruning LoRA and the LLM. \textbf{LoRA-drop}\cite{lora-drop} uses the output of LoRA at each layer to evaluate the importance of parameters and prune the unimportant parameters. By contrast, \textbf{LoRAPrune}\cite{loraprune} jointly pruning LoRA matrices and the LLM parameters based on LoRA’s gradients.
Besides, we can also use LoRA to support parameters pruning for LLMs~\cite{lorashear,zhu2023parameter}.

\subsubsection{Parameter Sharing}
\label{subsubsec:parameter-sharing}

Parameter-sharing methods reduce the number of parameters by sharing parameters across different layers or modules of LLMs. \textbf{VeRA}\cite{vera} and \textbf{VB-LoRA}\cite{vb-lora} are two representative parameter-sharing methods for LoRA. Specifically, VeRA proposes to share a pair of frozen random matrices across all layers and conduct layer-wise adaptation with ``scaling vectors''. By contrast, VB-LoRA proposes a ``divide-and-share'' paradigm, which divides LoRA’s low-rank decomposition by a rank-one decomposition and achieves global sharing based on an admixture model.
Instead of sharing parameters in the original parameter space, \textbf{FourierFT}~\cite{gao2024parameter} converts the incremental matrix $\Delta{W}$ into the spatial domain using Fourier transform. It shares spectral entries across all layers and only learns its sparse spectral coefficients for each layer, thus reducing the number of trainable parameters.
% 
% Parameter-sharing methods reduce the number of parameters that need to be updated and stored during fine-tuning by sharing parameters across different layers or modules, thus optimizing memory and computation costs. Specifically, VeRA and VB-LoRA \cite{vera, vb-lora} provide two distinct implementations in this regard. 

% VeRA\cite{vera} freezes a pair of randomly initialized matrices and shares these matrices across all LoRA layers. It introduces trainable scaling vectors for layer-wise adaptation. The trained scaling vectors and low-rank matrices can be merged into the original weights, eliminating additional inference latency. VB-LoRA\cite{vb-lora} proposes a "divide-and-share" paradigm, significantly reducing the number of trainable parameters in LoRA. Specifically, VB-LoRA divides the low-rank decomposition vectors of LoRA into sub-vectors and uses a globally shared vector bank to combine these sub-vectors. Each sub-vector is selected and combined from several basic elements in the vector bank through a top-k mixture module. This approach allows for global parameter sharing, significantly reducing storage and transmission costs while maintaining model performance.

\subsection{Parameter Quantization}
\label{subsec:parameter-quantization}

Quantization, which reduces the bit width of parameters (e.g., from 32-bit floats to 4-bit integers), can be used to reduce the memory and computational cost of LoRA. Existing quantization-aware LoRA methods consist of post-training quantization (PTQ)-based methods and quantization-aware training (QAT)-based methods\cite{L4q}.

% Due to the high memory and computational demands of LLMs, methods combining quantization and LoRA fine-tuning are gaining popularity. Quantization reduces the model's weights from high precision (e.g., 32-bit floats) to low precision (e.g., 8-bit or 4-bit integers). This significantly decreases computational complexity and memory usage, enhancing inference efficiency and deployment convenience. However, quantized models often experience a decline in performance compared to fully fine-tuned models. To bridge this performance gap, LoRA fine-tuning can be applied, enabling quantized models to achieve higher performance. Current methods for adapting LoRA to quantized models fall into two main categories: (1) fine-tuning only the LoRA parameters, and (2) simultaneously optimizing both quantization and LoRA parameters.

\subsubsection{PTQ-based methods}
\label{subsubsec:ptq-methods}

In PTQ-based methods, we first quantize an LLM and then fine-tune the quantized model, namely quantization and fine-tuning are sequentially conducted. \textbf{QLoRA}~\cite{QLoRA} is the first PTQ-based quantization-aware LoRA method. In the fine-tuning stage, it first quantizes an LLM to 4 bits and then fine-tunes a LoRA module on it with a higher precision, such as BFloat16 or Float16. In the inference stage, it dequantizes the LLM to the same precision as LoRA and then adds the LoRA updates to the LLM. 

Although QLoRA can significantly reduce memory cost for fine-tuning, it does not bring benefits for inference, because it requires dequantizing the LLM to high precision again. To solve this problem, \textbf{QA-LoRA}~\cite{QA-LoRA} is proposed to reduce memory cost for both the fine-tuning and inference stages. QA-LoRA uses group-wise operators to balance the degrees of freedom of the LLM quantization and fine-tuning, which enables it to obtain a LoRA module having identical precision with the quantized LLM.
Thus, it can perform inference without dequantization.

\subsubsection{QAT-based methods}
\label{subsubsec:qat-methods}

In QAT-based methods, we jointly quantize and fine-tune an LLM, namely quantization and fine-tuning are simultaneously conducted. These methods can alleviate the quantization discrepancies observed in PTQ-based methods. To address the quantization discrepancy of QLoRA, \textbf{LoftQ}~\cite{LoftQ} alternatively applies quantization and low-rank approximation during fine-tuning to minimize the quantization error. However, \textbf{ApiQ}~\cite{ApiQ}  points out that LoftQ ignores the error propagation across layers and proposes activation-preserved initialization to avoid error propagation. Besides, \textbf{L4Q}~\cite{L4q} is another QAT-based method that has an advanced layer design.

\subsection{Parallel LoRA Computing Frameworks}
\label{subsec:parallel_lora_computing}

LoRA's parameter-efficient nature enables us to fine-tune or infer multiple modules on a single GPU or a GPU cluster, which can save computational resources and improve the efficiency of LoRA. This section introduces the parallel fine-tuning and parallel inference frameworks, respectively.

% to fine-tune and inference multiple LoRA modules on a single GPU or a GPU cluster.

% The scale and complexity of existing models have brought significant computational and storage challenges. To efficiently train and deploy these models in a multi-tenant environment, LoRA service systems need to enhance performance through a series of technical strategies. Existing methods design two stages of optimization: training and inference.

\subsubsection{Parallel Fine-tuning}
\label{subsubsec:parallel_ft}

Parallelly fine-tuning multiple LoRA modules on a single GPU can reduce GPU memory usage and improve computation efficiency. \textbf{ASPEN}~\cite{aspen} proposes a high-throughput parallel finetuning framework for LoRA, which consists of a BatchFusion approach and an adaptive job scheduling algorithm. Specifically, the BatchFusion approach supports parallelly fine-tuning multiple LoRA modules on a shared LLM by fusing multiple input batches into a single batch, while the adaptive job scheduling algorithm allocates computation resources to the fine-tuning jobs.

% In the training stage, optimizing the LoRA service system is crucial for improving training efficiency and resource utilization. For example, ASPEN\cite{aspen} can efficiently train multiple LoRA tasks on a single GPU. It parallelizes the training process of multiple LoRA fine-tuning tasks through BatchFusion. Additionally, ASPEN introduces an adaptive job scheduling algorithm. This algorithm fine-tunes the scheduling based on job performance metrics to maximize system efficiency. 
% Another framework that optimizes the training phase is JORA\cite{jora}. The JORA framework optimizes the fine-tuning process of Llama-2 models by using LoRA. It uses JAX's just-in-time (JIT) compilation and tensor-sharding techniques to achieve efficient training of LoRA parameters.

\subsubsection{Parallel Inference}
\label{subsubsec:parallel-inference}

Parallel inference framework for LoRA can not only improve the computational efficiency but also support the needs of multi-tenant service. \textbf{Punica}\cite{punica} uses a new CUDA kernel design to batch GPU operations for different LoRA modules. Based on Punica, \textbf{S-LoRA}~\cite{s-lora} further optimizes the parallel inference framework by introducing a unified paging mechanism and a new tensor parallelism strategy, which enables the service of thousands of concurrent LoRA modules. Then, based on Punica and S-LoRA, \textbf{CARASERVE}~\cite{caraserve} reduces the cold-start overhead and further improves the service efficiency and SLO (service-level objective) attainment rates by CPU-GPU cooperation and rank-aware scheduling.

\section{LoRA for Federated Learning}
\label{sec:LoRA_Federate_Learning}
When adapting LLMs to vertical domains such as medicine and finance, the available training data can be privately owned by multiple clients. In this scenario, the training data is not centralized, and we have to fine-tune LLMs while keeping the data localized, namely federated learning. In federated learning, the clients typically compute weight updates locally and then share these updates with others to globally update the LLM. It brings both communication and computation costs for the clients. Fortunately, LoRA is parameter efficient and pluggable, which can reduce communication costs and lower computational resource requirements. LoRA can enhance the overall efficiency and scalability of federated learning. 

\begin{figure*}[ht]
    \centering
    \includegraphics[width=1.0\linewidth]{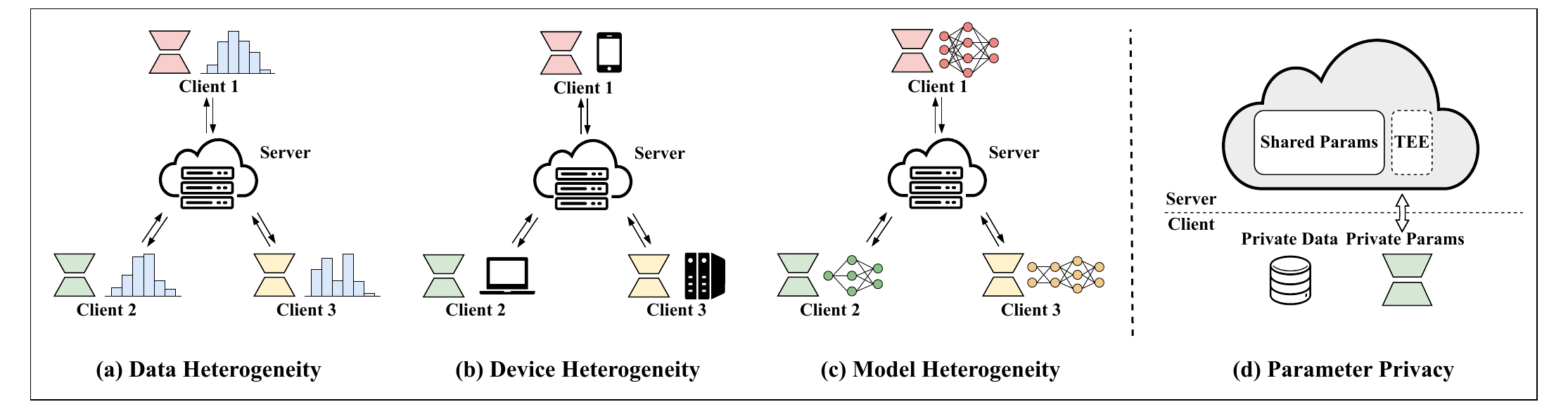}
    \caption{An illustration of LoRA for federated learning.}
    \label{fig:lora_federate}
\end{figure*}

However, adopting LoRA in federated learning is not trivial for federated learning faces challenges such as data heterogeneity, device heterogeneity, and model heterogeneity.
% 
 % This section introduces each category of methods, as illustrated in Fig.~\ref{fig:lora_computing}.
% 
To address these issues, recent studies have designed various methods for LoRA to meet the diverse needs of federated learning, as shown in Fig.~\ref{fig:lora_federate}. Additionally, as a localized parameter component, LoRA's pluggable nature allows it to support parameter privacy protection in federated learning. 
% 
% Federated learning is a secure distributed machine learning framework. It uses private data on edge devices to train a global model while keeping the data localized. However, frequent communication between devices results in significant cost. LoRA approximates model parameter changes with low-rank matrices. It reduces communication cost and lowers computational resource requirement, enhancing the overall efficiency and scalability of the system.
% 
% Federated learning faces challenges such as data heterogeneity, device heterogeneity, and model heterogeneity. To address these issues, recent studies have designed various methods for LoRA to meet diverse need of federated learning. Additionally, as a personalized parameter component, LoRA's modular nature allows it to support parameter privacy protection in federated learning.

\subsection{Data Heterogeneity}
\label{subsec:data_hetero}
Data heterogeneity refers to differences in data distribution across clients. In federated learning, different clients usually have different data distributions. The inconsistency in data distribution affects the overall performance of the model. Research reveals that in federated learning, as user data becomes more diverse, the performance gap between LoRA and full fine-tuning widens~\cite{slora}. To address this issue, researchers have proposed several improvement methods.

\textbf{SLoRA}~\cite{slora} introduces a data-driven initialization method for LoRA. It first performs sparse federated fine-tuning before applying LoRA and then performs SVD to decompose the accumulated gradient updates into low-rank matrices for LoRA initialization. The goal is to enable the LoRA modules to better adapt to the data distribution of each client, thereby integrating these heterogeneous data characteristics into the global model more effectively. \textbf{FeDeRA}~\cite{FeDeRA} uses a simpler initialization method. It directly applies SVD to pre-trained weights to initialize LoRA. Retaining the principal components of the pre-trained weights aligns the direction and magnitude of weight updates across different clients to handle data heterogeneity. Additionally, \textbf{FFA-LoRA}~\cite{FFA-LoRA} freezes one low-rank matrix and fine-tunes only the other. This reduces inconsistency during server aggregation of LoRA gradients, alleviating the optimization instability caused by non-IID data. 

\subsection{Device Heterogeneity}
\label{subsec:device_hetero}

Device heterogeneity refers to the differences in hardware capabilities, and network connectivity among clients participating in federated learning. Traditional federated learning methods often encounter the ``buckets effect'', implying that the system's overall performance is limited by the capability of the least powerful client. Specifically, these methods use the smallest LoRA rank to accommodate all clients, which prevents many resource-rich clients from fully utilizing their potential.

To address this issue, a dynamic parameter allocation strategy can be adopted. \textbf{FedMS}~\cite{FedMS} dynamically adjusts the number of activated LoRA matrices based on the real-time computational resources of clients. \textbf{FlexLoRA}~\cite{FlexLoRA} uses a dynamic parameter allocation strategy. It adjusts the LoRA rank and redistributes the SVD components of the global LoRA weights based on resource constraints. Similarly, \textbf{HETLORA}~\cite{HETLORA} assigns different ranks for different clients. However, it performs weighted aggregation according to the sparsity of the updates from different clients, balancing update information better than simple aggregation.

\subsection{Model Heterogeneity}
\label{subsec:model_hetero}
Model heterogeneity indicates differences in model structures among clients. In traditional federated learning, clients use local models with the same architecture, allowing their parameters to be aggregated into a global model on the server. However, in practice, clients may prefer unique local model architectures due to personal needs and often do not want to disclose model details. Thus, it is necessary to transfer knowledge between heterogeneous models without sharing private data or revealing local model structures~\cite{Heterogeneous_FL}.

Previous work has used knowledge distillation, model ensembling, and mutual learning to address model heterogeneity. However, these methods have limitations, such as reliance on public datasets, additional communication costs and poor local model performance. To avoid these limitations, \textbf{pFedLoRA}~\cite{pFedLoRA} uses LoRA as a carrier of both global and local knowledge. It adopts an iterative training strategy to facilitate knowledge transfer and integration, enabling knowledge sharing among heterogeneous models across different clients.

\subsection{Parameter Privacy}
\label{subsec:param_privacy}
In federated learning, protecting client-specific parameters is crucial because ensuring the privacy of these parameters also indirectly safeguards client data privacy. As a modular approach to adjusting personalized parameters, LoRA can be effectively integrated into federated learning systems to achieve parameter privacy protection.

Literature~\cite{TEE} proposes a secure distributed language model training framework based on model slicing. They deploy LoRA in a Trusted Execution Environment (TEE) and use OTP encryption to transmit features between the GPU and TEE, protecting model parameter privacy. \textbf{PrivateLoRA}~\cite{PrivateLoRA} introduces a distributed system based on LoRA. It adds a square matrix $M$ between low-rank matrices $A$ and $B$. The non-trainable matrices A and B, along with most of the pre-trained weights, are deployed on the global server to enhance computation. Meanwhile, the trainable matrix $M$ is stored on the client as personalized parameters, thus ensuring parameter privacy protection. 

Furthermore, recent works have integrated differential privacy (DP) techniques with LoRA in federated learning to enhance data privacy. \textbf{DP-LoRA}~\cite{DP-LoRA} ensures differential privacy by adding Gaussian noise to LoRA's weight updates during the update process. This approach maintains privacy and improves communication efficiency. To solve the noise amplification when applying differential privacy in LoRA, \textbf{FFA-LoRA}~\cite{FFA-LoRA} fixes the matrix $A$, avoiding the local semi-quadratic structure and enhancing robustness and performance.

\section{Applications of LoRA}
\label{sec:Applications}
In the rapidly evolving field of deep learning, LoRA has become widely used due to its unique advantages. Researchers utilize LoRA to fine-tune pre-trained models for various downstream tasks, reducing computational resource requirements while enhancing performance. LoRA's strong adaptability and efficiency have significantly improved various applications. In this section, we will introduce LoRA's applications in the following scenarios: (1) language tasks; (2) vision tasks; (3) multimodal tasks.
\subsection{Language Tasks}
\label{subsec:language_task}
Recently, the rapid development of pre-trained language models, especially LLMs, is revolutionizing the approach to language tasks due to their outstanding performance. However, these pre-trained models are trained on a large amount of general data and still require further fine-tuning on task-specific data to adapt to downstream tasks. Therefore, it is natural to use LoRA to fine-tune these pre-trained language models, as it reduces computational resource requirements. We mainly focus on some representative downstream tasks, which include traditional NLP tasks, code tasks, model alignment and vertical domain tasks.

\subsubsection{Traditional NLP Tasks}
Given the strong instruction-following and contextual understanding abilities of LLMs, some researches apply LoRA to fine-tune these models for traditional NLP tasks. For example, LoRA is widely adopted in LLaMA for various tasks, such as emotion recognition \cite{zhang2024dialoguellm}, text classification\cite{li2023label} and role recognition \cite{Bornheim_2024}. \textbf{AutoRE}~\cite{lilong2024autore} applies QLoRA to three document-level relation extraction tasks, achieving great performance on different LLMs. Some studies\cite{alves2023steering, zheng2024fine, mujadia2023assessing} leverage LoRA from different perspectives to enhance the model's capability in machine translation tasks. Additionally, LoRA can also improve the performance of models like BERT and T5 for text understanding tasks \cite{zhang2024personalized,liu2024tuning}.

\subsubsection{Code Tasks}
Some researchs apply LoRA to improve model performance in various code-related tasks. For example, BERT-style models fine-tuned with LoRA are suitable for code-change-related tasks, specifically in Just-In-Time defect prediction (JIT-DP)\cite{liu2024delving,guo2024empirical}. Similarly, training CodeT5 and PLBART with LoRA can enhance their adaptability for code summarization and code clone detection\cite{ayupov2022parameter}. As for the decoder-only model, \textbf{RepairLLaMA}\cite{silva2023repairllama} uses LoRA to fine-tune Llama for automated program repair (APR), while WizardCoder-15B is fine-tuned with LoRA for Text-to-SQL task\cite{roberson2024analyzing}. Additionally, \textbf{SteloCoder}\cite{pan2023stelocoder}, a fine-tuned version of StarCoder, is designed for multi-language to Python code translation.

\subsubsection{Model Alignment Tasks}
Model alignment tasks focus on adjusting a machine learning model to align with human values and intentions, often using techniques like Reinforcement Learning from Human Feedback (RLHF). To reduce memory requirements of RLHF, some studies use LoRA to fine-tune the reward model and policy model\cite{sidahmed2024perl,santacroce2023efficient,sun2023exploring}. Furthermore, other works improve reward models by integrating multiple LoRA adapters. For example, \textbf{DMoERM}\cite{quan2024dmoerm} combines MoE with LoRA, routing model inputs to multiple LoRA experts while another work\cite{zhang2024improving} proposes a LoRA-based ensemble method as well. The integration can also benefit the quantification of uncertainty in reward models\cite{zhai2023uncertaintypenalized}. Besides, literature\cite{yang2024bayesian} applies \textbf{Laplace-LoRA}\cite{yang2023bayesian} to train Bayesian reward models, which mitigates reward overoptimization in best-of-n sampling.

\subsubsection{Vertical Domain Tasks}
LLMs often perform suboptimally in vertical domains, requiring fine-tuning with domain-specific expertise. Some works apply LoRA to improve the performance of LLMs on domain-specific tasks. For example, some studies fine-tune LLMs on medical datasets with LoRA to adapt them to the medical domain\cite{tran2024bioinstruct,gema2024parameterefficient,toma2023clinical}. Additionally, other studies improve medical tasks like clinical dialogue summarization\cite{suri2023suryakiran}, assertion detection\cite{ji2024assertion} and medical QA tasks\cite{wang2023ivygpt,bhatti2023sm70}. Similarly, several studies fine-tune LLMs with LoRA on financial data to solve tasks such as financial news analytics and sentiment classification\cite{konstantinidis2024finllama, pavlyshenko2023financial,liu2023fingpt,li2024ra}. Besides, LoRA can also be used to enhance the performance in database tasks like query rewrite and index tuning\cite{zhou2024db}.

\subsection{Vision Tasks}
In vision tasks, LoRA is primarily applied to image generation and image segmentation, significantly improving training efficiency and optimizing model performance.
\label{subsec:vision_task}
\subsubsection{Image Generation}
Image generation tasks hold significant importance in the field of computer vision. In recent years, diffusion model have demonstrated exceptional performance in image generation tasks. LoRA is widely used in diffusion models to address various image generation tasks while reducing computational resources. Some works use LoRA to fine-tune diffusion models for image style transfer\cite{li2024diffstyler,frenkel2024implicit,liu2023facechain,liao2023calliffusion,shrestha2023style}, while others apply it to text-to-image generation\cite{li2024blockwise,kong2024omg,shi2023space,jin2023generating,wang2023customizing}.

Furthermore, researchers have designed several LoRA-based methods to improve image generation quality. For instance, \textbf{Smooth Diffusion}\cite{guo2023smooth} uses LoRA to achieve smoothness in the latent space, leading to better performance in various image generation and editing tasks. \textbf{ResAdapter}\cite{cheng2024resadapter} employs LoRA to learn resolution priors, adjusting the receptive fields of convolutional layers to dynamical resolution. Additionally, to specifically enhance text-to-image quality, \textbf{STAMINA}\cite{smith2024continual} uses LoRA to fine-tune diffusion models for longer concept sequences. \textbf{DreamSync}\cite{sun2023dreamsync} and \textbf{StyleAdapter}\cite{wang2023styleadapter} use LoRA to improve text fidelity and image quality. \textbf{Mix-of-Show}\cite{mix-of-show} captures out-of-domain information with LoRA weights to combine multiple customized concepts with high fidelity, reducing concept conflicts. Other studies combine LoRA with model distillation to accelerate image generation\cite{luo2023lcmlora,golnari2023loraenhanced}. Moreover, LoRA can also be applied to video generation\cite{ren2024customizeavideo,deng2024dragvideo,yang2023rerender,khandelwal2023infusion,blattmann2023stable,guo2023animatediff} and 3D generation tasks\cite{huang2024dreamcontrol,ma2023xdreamer,yu2023boosting3d,yoo2024plausible,zhang2024dragtex}.

\subsubsection{Image Segmentation}
Image segmentation is a significant challenge in computer vision, aiming to divide an image into multiple meaningful regions or objects. To address this, SAM has been proposed as a foundational model for image segmentation and demonstrated superior generalization ability. To further enhance its performance in specific vertical domains, many studies utilize LoRA to fine-tune it. For instance, in license plate detection, \textbf{SamLP}\cite{ding2024samlp} utilizes LoRA to adapt SAM for efficient segmentation of license plates. In structural damage detection, literature\cite{ye2024sambased} fine-tunes SAM's encoder using LoRA for instance segmentation task. In the medical domain, many studies also apply LoRA to fine-tune SAM for a variety of tasks, including nuclei segmentation\cite{na2024segment}, OCTA image segmentation\cite{chen2024samocta}, brain tumor segmentation\cite{feng2023cheap}, organ segmentation\cite{zhang2023customized}, and surgical instrument segmentation\cite{wang2023sam}. Additionally, some studies use LoRA to fine-tune Vision Transformer (ViT) for visual tracking\cite{lin2024tracking} and face forgery detection\cite{kong2023enhancing}.

\subsection{Multimodal Tasks}
\label{subsec:multimodal_task}
Multimodal Large Language Models (MLLMs) aim to integrate text with various modalities such as audio, image and video, which enable cross-modal understanding and reasoning through a unified embedding space. The success of LoRA in both NLP and vision tasks has sparked considerable interest in applying them to MLLMs.

In MLLMs, LoRA can not only improve training efficiency but also facilitate effective modality alignment. In audio-text tasks, \textbf{SALM}\cite{chen2023salm} comprises LoRA layers, a frozen text-based LLM, an audio encoder and a modality adapter to handle speech inputs and corresponding task instructions. For image-text tasks, \textbf{InternLM-XComposer2}\cite{dong2024internlmxcomposer2} achieves modality alignment by applying LoRA to image tokens, \textbf{mPLUG-Owl}\cite{ye2024mplugowl} freezes the visual module while jointly fine-tuning LoRA and abstractor of the text module, and \textbf{CoLLaVO}\cite{lee2024collavo} employs QLoRA to preserve object-level image understanding. In the realm of video-text tasks, \textbf{VSP-LLM}\cite{yeo2024visual} fine-tunes the text module with QLoRA for visual speech processing, \textbf{MolCA}\cite{liu2024molca} uses LoRA to understand 2D molecular graphs and text, while \textbf{TPLLM}\cite{ren2024tpllm} employs LoRA for efficient traffic prediction by integrating sequence and spatial features. These applications demonstrate the versatility and power of LoRA in MLLMs tasks.

\section{Conclusion and Future Direction}
\label{sec:Conclusion}
In this survey, the recent progress of LoRA have been systematically reviewed from the perspective of downstream adaptation improving, cross-task generalization, efficiency improving, federated learning and applications. From this review, we can find that LoRA is parameter efficient, pluggable, compatible and easy to achieve cross-task generalization, which enables it to be one of the most important technology for LLMs applications. Recent progress further boosts the generalization and efficiency of LoRA, and stimulate its potential to be used in more scenarios. Here, we list three future directions where LoRA will be indispensable.

\subsection{LoRA for GaaS}

In Generative-as-a-Service (GaaS), cloud-based platforms provide users with generative artificial intelligence (AGI) services. GaaS enables users enjoy AGI without deploying local computational resources. For the users' needs are diverse, it is necessary to provides various functions for GaaS. To implement the various functions, we can construct a LoRA module for each function. The pramameter efficiency and plugability of LoRA can facilitate efficient functions' construction and execution. Besides, the services on  GaaS platforms can change rapidly alonging time. To follow the changes, we can train new LoRA modules that initialized by combination of previous LoRA modules. The cross-task generalization ability of LoRA can facilitate fast adaption to service updations.

\subsection{LoRA for Continued Pre-training}

In continued pre-training, a foundation model is continuely trained with unlabeled user data to adapt the model to specific domains. Typically, the self-supervised training objective is same with that for pre-training, and the learning rate is much smaller than than for pre-training. Continued pre-training is a important stage for constructing vertical domain LLMs. However, it is highly computational expensive, which impedes the development of vertical domain LLMs, especailly for the organizations with limited computational resources. Enhancing LoRA for continued pre-training and reducing its computational cost is worth to explored.

\subsection{LoRA for Autonomous Agents}

In LLM-based autonomous agents, the agents are assigned with specific roles. Based the roles and environment,  agents make actions to response users' or other agents' request. The actions can be made based on self-knowledge or tools that designed for domain-specific tasks. The request and the actions are stored in memory to support the future requests.

In the current agents, the roles are typically assigned by prompts; however, prompt may cannot give a comprehensive discription of the role when the role is complex and the  number of related data is large. Assiging roles with LoRA modules training from data related to the roles  can be a better choice. Furthermore, the tools for agent can be LoRA modules. Besides, the memory usually augments the agents with retrieval augmented generation (RAG); however, due to the input token limitation and the short-comings of in-context learning, the RAG-based support may be less effective. By contrast, we can use LoRA-based continual learning to construct memory modules, which can solve the problem of RAG. Therefore, LoRA-driven agents   are worth to explore.

\begin{acknowledgement}
This work was supported in part by the NSFC under Grants No. (62025206, 62302436, U23A20296), Zhejiang Province's "Lingyan" R\&D Project under Grant No. 2024C01259, and Ningbo Science and Technology Special Projects under Grant No. 2023Z212. Yunjun Gao is the corresponding author of this work.
\end{acknowledgement}

% \section*{Appendixes~(if needed)}

% \subsection*{Appendix A}

% \subsection*{Appendix B}

\bibliographystyle{fcs}
\bibliography{main}

% \vspace*{-15ex}
\begin{biography}{authors/myr}
{Yuren Mao} received
his PhD degree under the supervision of Prof. Xuemin Lin in computer science from University of New South Wales, Australia in 2022. He is currently an assistant professor with the School of Software Technology, Zhejiang University, China. His current research interests include Large Language Models and its applications in Data Intelligence.
\end{biography}

% \vspace*{-8ex}
\begin{biography}{authors/yhge}
{Yuhang Ge} is currently working toward his PhD degree in the School of Software Technology at Zhejiang University, China. His research interests include Large Language Models and Data Management.
\end{biography}

\begin{biography}{authors/yjf}
{Yijiang Fan} is currently studying as a master's student in the School of Software Technology at Zhejiang University, China. His research interests include Large Language Models and collaborative inference.
\end{biography}

% \vspace*{3ex}
\begin{biography}{authors/wyxu}
{Wenyi Xu} is currently studying as a master's student in the School of Software Technology at Zhejiang University, China. His research interests include Multimodal Large Models and RAG.
% Retrieval-Augmented Generation.
\end{biography}

% \vspace*{3ex}
\begin{biography}{authors/ymi}
{Yu Mi} is currently studying as a master's student in the School of Software Technology at Zhejiang University, China. Her research interests include Large Language Models and AI for science.
\end{biography}

\begin{biography}{authors/hzh}
{Zhonghao Hu} is currently studying as a master's student in the School of Software Technology at Zhejiang University, China. His research interests include Large Language Models and data discovery.
\end{biography}

% \vspace*{-35ex}
\begin{biography}{authors/yjgao}
{Yunjun Gao} received the PhD degree in computer science from Zhejiang University, China, in 2008. He is currently a professor in the College of Computer Science  and Technology, Zhejiang University, China. His research interests include Database, Big Data Management and Analytics, and AI interaction with DB technology.
\end{biography}

\end{sloppypar}
\end{document}